\documentclass[10pt,twocolumn,letterpaper]{article}

\usepackage{cvpr}              
\usepackage[accsupp]{axessibility}  

\DeclareUnicodeCharacter{751F}{~}
\DeclareUnicodeCharacter{614B}{~}
\DeclareUnicodeCharacter{5B66}{~}
\DeclareUnicodeCharacter{7684}{~}
\DeclareUnicodeCharacter{8996}{~}
\DeclareUnicodeCharacter{899A}{~}
\DeclareUnicodeCharacter{8AD6}{~}

\usepackage[utf8]{inputenc} 
\usepackage[T1]{fontenc}    
\usepackage{url}            
\usepackage{booktabs}       
\usepackage{amsfonts}       
\usepackage{nicefrac}       
\usepackage{microtype}      
\usepackage{makecell}
\usepackage{graphicx}
\usepackage{amsmath}
\usepackage{bm}
\usepackage{multirow} 
\usepackage{stfloats}
\usepackage{diagbox}
\usepackage{amsthm}

\usepackage{color}
\usepackage[table,xcdraw]{xcolor}
\usepackage{pgfplots}
\usepackage{pgfplotstable}
\usepackage{geometry}

\pgfplotsset{compat=1.18}
\usepackage{caption}
\usepackage[font=small, labelfont=bf]{caption}

\usepackage{lipsum}

\usepackage[pagebackref,breaklinks,colorlinks,linkcolor=red,urlcolor=red,citecolor=blue,bookmarks=false]{hyperref}

\hypersetup{pdfstartview={FitH}}

\def\paperID{42566} 
\def\confName{CVPR}
\def\confYear{2026}
\author{Nghia Vu$^{\dagger, 2}$, Tuong Do$^{\dagger, 1,2,3}$, Khang Nguyen$^{4}$, Baoru Huang$^{1,7,*}$, Nhat Le$^{5}$, Binh Xuan Nguyen$^{2}$,\\ Erman Tjiputra$^{2}$, Quang D. Tran$^{1,2}$, Ravi Prakash$^{6}$, Te-Chuan Chiu$^{3}$, Anh Nguyen$^{1}$\\
{\footnotesize $^{1}$University of Liverpool, UK} \ \ \ {\footnotesize $^{2}$AIOZ Ltd., Singapore}\ \ \ {\footnotesize $^{3}$National Tsing Hua University, Taiwan}\\ 
{\footnotesize $^{4}$MBZUAI}\ \ {\footnotesize $^{5}$University of Western Australia}\ \ \ {\footnotesize $^{6}$Indian Institute of Science} \ \ \
{\footnotesize $^{7}$NVIDIA}\\
\footnotesize \href{https://aioz-ai.github.io/AffordMatcher/}{https://aioz-ai.github.io/AffordMatcher/}
}
\vspace{-7pt}

\begin{document}


\title{\textit{AffordMatcher}: Affordance Learning in 3D Scenes from Visual Signifiers}

\twocolumn[{%
\renewcommand\twocolumn[1][]{#1}%
\maketitle
\begin{center}
    \centering
    \vspace{-22pt}
    \captionsetup{type=figure}
    \includegraphics[width=1.00\linewidth]{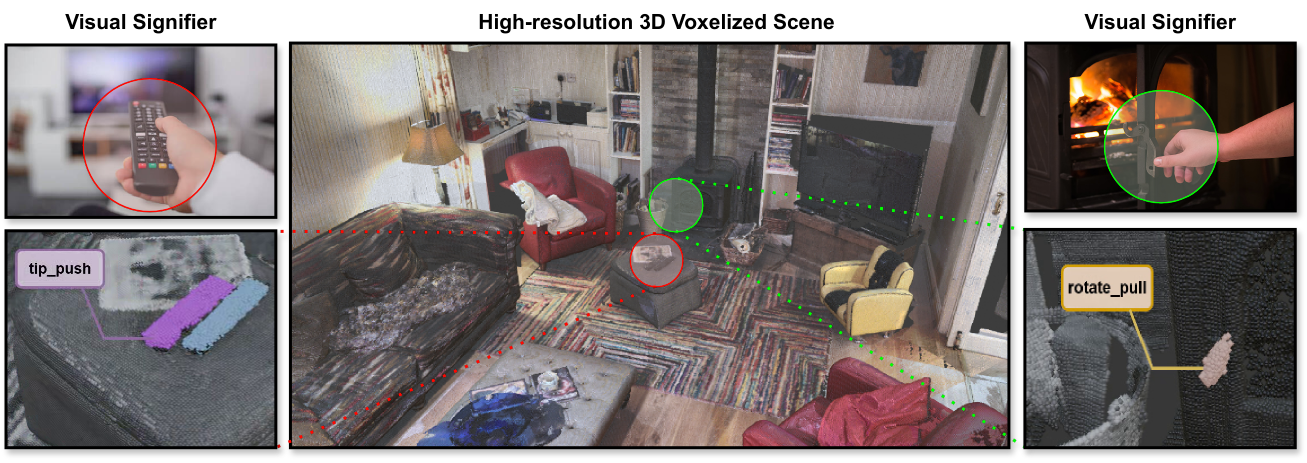}
    \vspace{-21pt}
    \captionof{figure}{\textbf{Overview of \textit{AffordMatcher}:} Detecting and localizing affordances in 3D voxelized scenes through visual signifiers entails semantic context drawn from RGB images. Given a scene representation and visual signifiers, \textit{AffordMatcher} can understand actionable commands, such as ``\textit{watch the television}'', ``\textit{push the tip}'', ``\textit{rotate pull}'', or ``\textit{open the chimney}'', and identify spatial affordances.}
    \label{fig:teaser}
\end{center}
}]
\vspace{-10ex}
\begin{abstract}
\vspace{-0ex}
Affordance learning is a complex challenge in many applications, where existing approaches primarily focus on the geometric structures, visual knowledge, and affordance labels of objects to determine interactable regions. However, extending this learning capability to a scene is significantly more complicated, as incorporating object- and scene-level semantics is not straightforward. 
In this work, we introduce \textit{AffordBridge}, a large-scale dataset with $291,637$ functional interaction annotations across $685$ high-resolution indoor scenes in the form of point clouds. Our affordance annotations are complemented by RGB images that are linked to the same instances within the scenes. Building upon our dataset, we propose \textit{AffordMatcher}, an affordance learning method that establishes coherent semantic correspondences between image-based and point cloud-based instances for keypoint matching, enabling a more precise identification of affordance regions based on cues, so-called visual signifiers.\renewcommand{\thefootnote}{}
\setcounter{footnote}{0}
\footnotetext{
$^{\dagger}$ equal contribution; * corresponding author
}\noindent Experimental results on our dataset demonstrate the effectiveness of our approach compared to other methods. 

\end{abstract}

\vspace{-20pt}
\section{Introduction}
\vspace{-3pt}
Humans interact with the environment as part of their daily routines. Analyzing these interactions can provide valuable insights and is highly beneficial for both humans and robots to perform meaningful actions in the given environment. To better understand the interaction between humans and the environment, Gibson introduced the concept of ``\textit{affordance}'', referring to ``\textit{opportunities for interaction}''~\cite{Gibson1979-GIBTEA}.  Yet, these opportunities need to be physically verifiable through successful actions to determine whether they are true \textit{signifiers}; one point was later discussed by Norman~\cite{norman2013design}. Therefore, learning about affordances requires not only predicting types of interactions but also correctly identifying specific points on objects that facilitate these human-object interactions. The concept of affordance from signifiers brings together perception and action, thus opening up applications in robotic manipulation~\cite{bahl2023affordances, xu2021affordance, wu2023learning, chung2026rethinking}, human-robot interaction~\cite{zeng2022robotic, braud2020robot}, visual navigation~\cite{lee2024affordance, NEURIPS2020_15825aee}, and augmented reality~\cite{raikwar2019cubevr, yoo2023augmenting}.

\begin{table*}[ht!]
    \resizebox{\linewidth}{!}{
    \setlength{\tabcolsep}{0.30em} 
    {\renewcommand{\arraystretch}{1.2} 
    \begin{tabular}{r|cc|cc|cccc}
        \toprule
        \multirow{2}{*}{\textbf{Dataset} / \textbf{Attribute}} & \multirow{2}{*}{\textbf{\begin{tabular}[c]{@{}c@{}}Total \\ Samples\end{tabular}}} & \multirow{2}{*}{\textbf{Environment}} & \multicolumn{2}{c|}{\textbf{Form of Interactions}} & \multirow{2}{*}{\textbf{\begin{tabular}[c]{@{}c@{}}Affordance \\ Annotation\end{tabular}}} & \multirow{2}{*}{\textbf{\begin{tabular}[c]{@{}c@{}}No. \\ Affordances\end{tabular}}} & \multirow{2}{*}{\textbf{\begin{tabular}[c]{@{}c@{}}No. \\ Categories\end{tabular}}} & \multirow{2}{*}{\textbf{\begin{tabular}[c]{@{}c@{}}No. Aff.\\ Actions\end{tabular}}} \\ \cline{4-5}
        &  &  & \multicolumn{1}{c|}{\textit{Implicit}} & \textit{Explicit} &  &  &  &  \\ \hline
        \rowcolor[HTML]{EFEFEF}EPIC-Aff~\cite{nagarajan2020ego} & 38,876 & 2D Images & \multicolumn{1}{c|}{--} & -- & 2D masks & -- & 304 & 43 \\ 
        AGD20k~\cite{luo2022learning} & 23,816 & 2D Images & \multicolumn{1}{c|}{--} & -- & 2D boxes & -- & 50 & 36 \\ 
        \rowcolor[HTML]{EFEFEF}PartNet~\cite{mo2019partnet} & 26,671 & 3D Objects & \multicolumn{1}{c|}{--} & -- & 3D masks & 573,585 & -- & 24 \\ 
        AffordPose~\cite{jian2023affordpose} & 641 & 3D Objects & \multicolumn{1}{c|}{--} & \begin{tabular}[c]{@{}c@{}}Grasp hand poses\end{tabular} & 3D masks & 26,712 & 13 & 8 \\ 
        \rowcolor[HTML]{EFEFEF}3DAffordanceNet~\cite{AffordanceNet18} & 22,949 & 3D Objects & \multicolumn{1}{c|}{--} & -- & 3D masks & 56,307 & 23 & 18 \\ 
        PIAD~\cite{yang2023grounding} & 7,012 & 3D Objects & \multicolumn{1}{c|}{\begin{tabular}[c]{@{}c@{}}Single interactions\end{tabular}} & -- & 3D masks & 7,012 & 23 & 17 \\
        \rowcolor[HTML]{EFEFEF}LASO~\cite{li2024laso} & 8,434 & 3D Objects & \multicolumn{1}{c|}{\begin{tabular}[c]{@{}c@{}}Commands on objects\end{tabular}} & -- & 3D masks & 19,751 & 23 & 17 \\ 
        Scenefun3D~\cite{delitzas2024scenefun3d} & -- & 3D Scenes & \multicolumn{1}{c|}{--} & \begin{tabular}[c]{@{}c@{}} Commands on scenes\end{tabular} & 3D masks & 14,279 & -- & 9 \\ 
        \rowcolor[HTML]{EFEFEF}PIADv2~\cite{shao2025great} & 38,889 & 3D Objects & \multicolumn{1}{c|}{\begin{tabular}[c]{@{}c@{}}Single interaction\end{tabular}} & -- & 3D masks & 38,889 & 43 & 24 \\ 
        AED~\cite{li2025learning} & -- & 2D Images & \multicolumn{1}{c|}{--} & \begin{tabular}[c]{@{}c@{}}-- \end{tabular} & 2D masks & -- & 13 & 8 \\ 
        \rowcolor[HTML]{EFEFEF}SeqAfford~\cite{Yu_2025_CVPR} & 18,371 & 3D Objects & \multicolumn{1}{c|}{\begin{tabular}[c]{@{}c@{}}Commands on objects\end{tabular}} & -- & 3D masks & 183,233 & 23 & 18 \\ 
        MIPA~\cite{gao2025learning} & 7,012 & 3D Scenes & \multicolumn{1}{c|}{Multiple interactions} & \begin{tabular}[c]{@{}c@{}} -- \end{tabular} & 3D masks & 7,012 & 23 & 17 \\ 
        \midrule
        \rowcolor[HTML]{EFEFEF}\textbf{\textit{AffordBridge} (Ours)} & 317,844 & 3D Scenes & \multicolumn{1}{c|}{\begin{tabular}[c]{@{}c@{}}Visual signifiers\end{tabular}} & \begin{tabular}[c]{@{}c@{}}Descriptions\end{tabular} & 3D masks & 291,637 & 157 & 61 \\ 
        \bottomrule
    \end{tabular}
    }}
    \vspace{-8pt}
    \caption{\textbf{Comparisons between \textit{AffordBridge} and other datasets:} Our dataset introduces a large-scale benchmark for spatial affordance identification from visual signifiers through both implicit and explicit human-object interactions. \textit{AffordBridge} contains $317,844$ high-resolution paired samples of 2D-3D representations across $685$ scenes. The interacted objects are annotated by 3D masks, yielding $291,637$ volumetric masks of interactable regions among $157$ object categories through $61$ actionable affordances.}
    \vspace{-16pt} 
    \label{tab:datasetComparison}
\end{table*}

Still, the concept of ``\textit{affordance}'' standalone is broad. Depending on the context, affordance learning is typically treated as a single-modality learner. Image-based affordance learning predicts the corresponding segmentation maps at the pixel level for intended actions~\cite{lueddecke2019context, roy2016multi,nguyen2016detecting, AffordanceNet18}. Meanwhile, spatial affordance learning segments desired affordance masks on object point clouds at the voxel level~\cite{deng20213d, Nguyen2023open,van2024open}. The share of the two mentioned approaches leverages text prompts to direct affordance learning. However, the fusion of these two modalities remains unclear~\cite{luo2023leverage, nguyen2024lightweight}. Through this, the lack of multimodal representation learning for affordance conveys the idea of imposing cross-modal learning.

In practice, many challenges coexist in localizing spatial affordance from visual signifiers. First, cross-modal representations require overcoming significant discrepancies in feature distributions between images and point clouds~\cite{Dharmasiri2024CrossModal}. Second, matching affordance localization in the 3D domain and affordance detection in image space under diverse actions across different scenes entails the dexterity of designing such learning models~\cite{Dharmasiri2024CrossModal}. Additionally, language instructions, such as ``\textit{press here}'' or ``\textit{rotate knob}'', are indeed semantically ambiguous without explicit geometric context~\cite{yang2023grounding}, not to mention that real‐world scans can be further degraded by noises and occlusions, which complicate purely geometry-based methods~\cite{zhang2023part}. Last but not least, most existing datasets lack paired RGB images with annotated 3D affordance regions tied to interaction cues, precluding learning models from end‐to‐end training and evaluation~\cite{Yu_2025_CVPR}. Without a unified solution to tackle these issues, creating an affordance bridge to localization affordances from visual signifiers is a utopian task. The central question is: ``\textit{How can we learn representations that match diverse spatial affordances across different scenes from visual signifiers?}''

To tackle this problem, we introduce a new benchmarking dataset, namely \textit{AffordBridge}, with RGB image–point cloud affordance annotations, and propose a visual‐guidance reasoning affordance method, so-called \textit{AffordMatcher}, that explicitly grounds visual signifiers into precise spatial affordances. As shown in Table~\ref{tab:datasetComparison}, our large‑scale benchmark has $317,844$ high‑resolution paired samples of 2D-3D representations in $685$ indoor scenes, resulting in $291,637$ volumetric masks of functionally interactive elements among $157$ object categories through $61$ actionable affordances. Building upon this dual modality, we develop our affordance learning model, \textit{AffordMatcher}, to semantically align keypoints in visual signifiers with those in point cloud instances, which accurately localize and segment interactable regions from both modalities, as shown in Fig.~\ref{fig:teaser}. To summarize, our contributions are twofold, as follows: 
\begin{itemize}
    \item \textbf{Affordance Dataset:} We introduce \textit{AffordBridge}, a large-scale dataset that annotates high-resolution point clouds and RGB images, alongside language-descriptive actions, for spatial affordance localization.
    \item \textbf{Affordance Learning:} We propose \textit{AffordMatcher} for effectively matching affordance regions in point clouds derived from RGB images.
\end{itemize}

\vspace{-13pt}
\section{Related Work}

\begin{figure*}[t]
    \centering
    \includegraphics[width=1.00\linewidth]{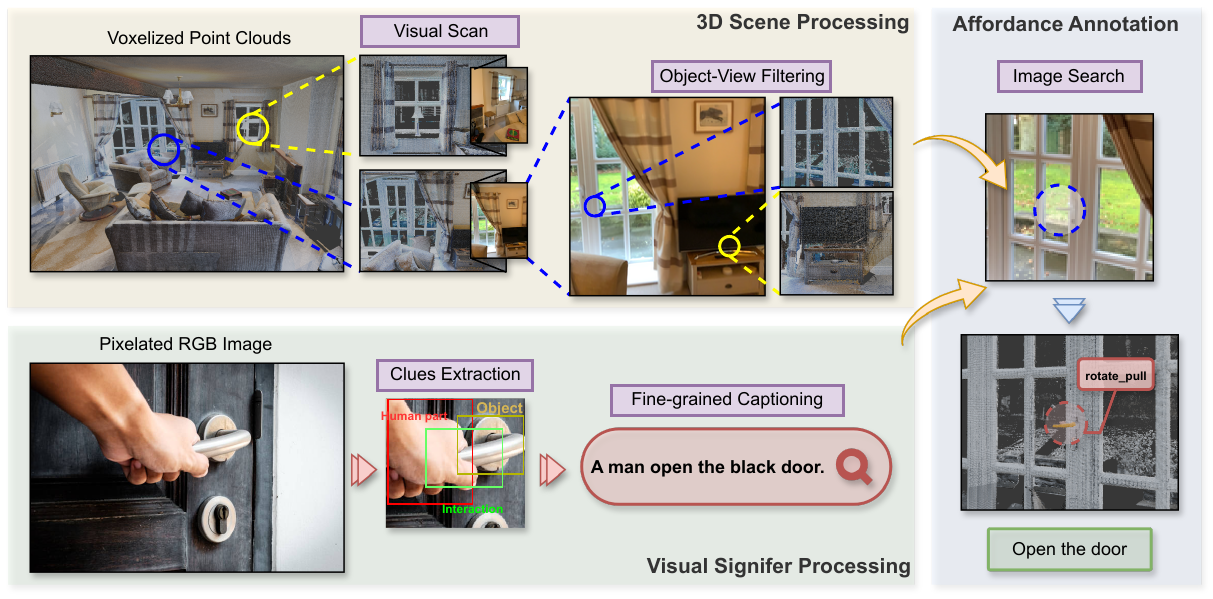}
    \vspace{-20pt}
    \caption{\textbf{Construction of the \textit{AffordBridge} dataset:} Our \textit{AffordBridge} dataset is built through a semi-supervised pipeline linking visual signifiers with 3D affordances. The building process includes (\textit{i}) 3D scene processing via voxelized point clouds with object-view filtering through visual scanning, (\textit{ii}) visual signifiers processing with human-object interaction extraction with fine-grained captioning, and (\textit{iii}) affordance annotation by matching key views to 3D instances for spatial action labeling.}
    \label{fig:dataset}
    \vspace{-16pt}
\end{figure*}
\vspace{-7pt}
\textbf{Affordance Datasets.} Existing affordance datasets primarily focus on annotating functional regions for human-object interactions, often at either the pixel or voxel level. Focusing images, EPIC-Aff~\cite{nagarajan2020ego}, IIT-AFF~\cite{nguyen2017object}, and AGD20k~\cite{luo2022learning} provide annotations with instance segmentation masks. Voxel-level annotated datasets, such as PartNet~\cite{mo2019partnet}, 3DAffordanceNet~\cite{AffordanceNet18}, and AffordPose~\cite{jian2023affordpose}, offer spatial masks for affordances; yet, they are all limited to distinct objects rather than full scenes. Recent works, PIAD~\cite{yang2023grounding}, LASO~\cite{li2024laso}, and SeqAfford~\cite{Yu_2025_CVPR}, incorporate implicit interactions through single commands or sequences of objects; meanwhile, Scenefun3D~\cite{delitzas2024scenefun3d} and MIPA~\cite{gao2025learning} extend to 3D scenes with fewer samples and limited affordance diversity, however. Most datasets emphasize object-level affordances (\textit{e.g.}, under $40,000$ samples and fewer than $25$ actions) that struggle to capture small functional details in complex scenes and are complicated by the integration of multimodal signifiers. To address these limitations, we propose a large-scale dataset of 3D scenes that incorporates both implicit and explicit descriptions of interactions among instance categories and actions for affordance learning from visual signifiers.

\noindent \textbf{Affordance Learning.} Prior studies on affordance learning focused on detecting affordance regions in images~\cite{AffordanceNet18, chen2023affordance, tang2023cotdet}, while recent research expands affordance learning into the spatial domain~\cite{ning2023where2explore, chen2024sugar}. 3D AffordanceNet~\cite{deng20213d} introduced the first benchmark dataset for learning affordances from 3D objects. Additionally, several methods incorporate additional information, such as images~\cite{yang2023grounding}, drone-related information~\cite{vu2026AeroScene}, or natural language instructions~\cite{li2024laso,engelmann2024opennerf}, to enhance the reasoning behind affordance regions. However, existing methods are mostly limited to point-level or object-level detection. Recent works have explored 3D indoor scene understanding guided by open-vocabulary~\cite{takmaz2023openmask3d, shi2024language, ding2023pla, nguyen2024open3dis}. Recently, SceneFun3D~\cite{delitzas2024scenefun3d} was proposed as a high-quality dataset for affordance understanding, featuring diverse natural language descriptions. Nevertheless, SceneFun3D~\cite{delitzas2024scenefun3d} only detects the affordances from the text prompt input. In this work, we focus on localizing spatial affordance from visual signifiers that provide human-object interactions.

\noindent \textbf{Semantic Correspondence.} Semantic matching focuses on identifying corresponding elements between the same instances in multiview settings. Traditional approaches achieve this by matching pixels or patches in given image pairs using fine-grained extracted features, which are thus matched through a correlation map using convolutional layers~\cite{li2020correspondence, min2020dhpf,do2024fine, sarlin2020superglue}, transformer networks~\cite{kim2022transformatcher, sun2021loftr, sun2023correspondence, sun2024pixel}, or another dedicated backbone~\cite{li2024sd4match, tang2023emergent} between views. Expanding beyond RGB images, modern approaches explore spatial feature matching. For example, DenseMatcher~\cite{zhu2024densematcher} combines the generalization capability of 2D foundation models with 3D geometric understanding to match textured 3D object pairs. 2D3D-MATR~\cite{li20232d3d} first establishes coarse correspondences between local patches in the point cloud and image view, then applies multi-scale patch matching to learn global contextual constraints. Here, our approach employs match-to-match attention to analyze cross-modal point cloud-image correspondences through affordances.

\vspace{-8pt}
\section{The \textit{AffordBridge} Dataset}
\vspace{-4pt}
In Fig.~\ref{fig:dataset}, we outline our three-stage annotation flow, which includes 3D scene processing (Sec.~\ref{sec:data_3d_proc}), visual signifier processing (Sec.~\ref{sec:data_2d_proc}), and affordance annotation (Sec.~\ref{sec:data_aff_annotate}). 

Let the colored scene point cloud be $\mathcal{P} = \{(p_{i}, f_{i})\}_{i=1}^{N}$, where $p_{i} \in \mathbb{R}^{3}$ denotes the 3D coordinates, and $f_{i} \in \mathbb{R}^{6}$ represents per-point features, including RGB colors and surface normals. From raw scans in~\cite{delitzas2024scenefun3d}, the point clouds are downsampled to $100,000$ points via voxelization with a voxel size of $0.05$ meters~\cite{mao2024denoising}, while preserving the details of scene representations and maintaining frugality for the affordance learning model. Thus, instance segmentation masks are applied to extract object regions within the scene~\cite{kolodiazhnyi2024oneformer3d}.
\vspace{-8pt}
\subsection{3D Scene Processing}
\vspace{-4pt}
\label{sec:data_3d_proc}
\textbf{Visual Scan.} Each colored scene point cloud is temporally aligned with the corresponding RGB video sequence $\mathcal{V} = \{ v_{k} \}_{k=1}^{K}$, where each frame $v_{k}$ represents a visual observation captured from a calibrated camera pose $\left[ R_{k} \mid t_{k} \right]$ using the camera's intrinsic matrix, denoted as $K_{c}$. Through SLAM-based trajectory estimation~\cite{zhu2022niceSLAM}, each frame $v_{k}$ is projected onto its associated 3D segment through depth alignment, ensuring geometric and temporal consistency along the trajectory. Multiview checks are used to remove occlusions and misalignments, yielding clean 2D-3D correspondences, denoted as $(P_{k}, v_{k})$, for downstream annotation. 

\noindent \textbf{Object-View Filtering.} Each visual scan may contain multiple objects with potential affordances. We detect candidate objects using MobileNet~\citep{howard2017mobilenets} and rank them by spatial location, scale, and contextual relevance. Each detected instance $v_{k}^{(l)}$ is therefore aligned with its 3D counterpart $P_{k}^{(l)}$, and inconsistent matches of approximately $15\%$ are manually discarded. To maintain reliability, we ensure inter-annotator agreements with a Kappa score higher than $0.75$~\cite{cohen1960coefficient} among three human experts reviewing the annotations.
\vspace{-4pt}
\subsection{Visual Signfier Processing}
\vspace{-4pt}
\label{sec:data_2d_proc}
\textbf{Annotation of Visual Signifiers.} We adopt interaction images from PIAD~\citep{yang2023grounding}, retaining only those that depict direct human-object contact. Each image is annotated with three bounding boxes $b = (b_{H}, b_{O}, b_{I})$ for human, object, and interaction regions, following the notation of MUREN~\citep{kim2023relational}. We refine the interaction box $b_{I}$ by inspecting model-predicted class scores to precisely localize the contact region. To further address potential ambiguities in visual signifiers, as static images may miss dynamics, we incorporate human pose estimation via OpenPose~\citep{cao2019openpose} to annotate keypoint-based hand-object contacts, enhancing the bounding boxes for humans, objects, and contact regions.

\noindent \textbf{Fine-grained Captioning.} 
Next, we generate fine-grained captions using visual signifiers as inputs. Specifically, using the Object Relation Transformer (ORT)~\citep{herdade2019image}, we fuse the three bounding boxes $b$ to produce coherent, templated descriptions. As shown in Fig. \ref{fig:dataset}, the caption ``\textit{A man opens the black door}'' describes the RGB image input. All captions are manually verified for semantic and spatial accuracy.
\vspace{-4pt}
\subsection{Affordance Annotation}
\vspace{-4pt}
\label{sec:data_aff_annotate}
To associate textual descriptions with corresponding scene views, we align image-text embeddings using CLIP encoders~\citep{radford2021learning} and retrieve the most relevant key-view by maximizing cosine similarity. The embedding space is refined through contrastive learning~\citep{oord2018representation} to achieve higher similarity between positive image-text pairs and vice versa. 

After filtering, each key-view $I_{i}$ is paired with an affordance action $a_{i}$ and its corresponding region within $\mathcal{P}$. We use a web-based annotation interface~\citep{dai2017scannet}, where annotators are able to identify the 3D instance segmentation mask $M_{i}$ that corresponds to the 2D view with the affordance mask $A_{i}$ for the affordance action $a_{i}$. The mapping between 2D affordances and 3D instances is $M_{i} = \arg \max_{M_{j} \in \mathcal{P}}\phi_{\text{sim}}\left(I_{i}, M_{j}\right)$,
where $\phi_{\text{sim}}(I_{i}, M_{j})$ measures the visual-geometric similarity between the key-view and the 3D object. To avoid one-to-many ambiguities, where a single visual signifier may correspond to multiple 3D instances, we allow multi-instance projection by retaining top-$3$ matches based on CLIP similarity scores. Subsequently, annotators verify and refine these candidates to obtain the most accurate correspondence. The resulting annotations yield object-level affordance masks embedded within full-scene geometry, with approximately $5\%$ of samples re-annotated to resolve ambiguous cases.

\begin{table}[h]
    \centering
    \vspace{-6pt}
    \resizebox{0.90\linewidth}{!}{
    \begin{tabular}{r|ccc|c}
        \toprule
        \diagbox{\textbf{Criteria}}{\textbf{Split}} & \textbf{Train} & \textbf{Validate} & \textbf{Test} & \textbf{Total} \\
        \midrule \midrule
        Visual Signifiers & $6,416$ & $1,974$ & $1,480$ & $9,870$ \\ 
        3D Scenes & $448$ & $138$ & $103$ & $689$ \\ 
        Affordance Areas & $189,564$ & $58,327$ & $43,746$ & $291,637$ \\ 
        Total Samples & 206.6K & 63.6K & 47.7K & 317.8K \\ 
        \bottomrule
    \end{tabular}
    }
    \vspace{-6pt}
    \caption{Train/validation/test split of the \textit{AffordBridge} dataset.}
    \label{tab:datasetSplit}
    \vspace{-14pt}
\end{table}
\vspace{-4pt}
\subsection{Dataset Statistics}
\label{sec:affordbridge_stats}
\vspace{-4pt}
Our \textit{AffordBridge} dataset is organized into training, validation, and test sets across three modalities: visual signifiers, 3D scenes, and affordance areas, as shown in Table~\ref{tab:datasetSplit}. The visual signifiers subset contains $9,870$ samples, while the 3D scenes subset includes $689$ samples. Affordance areas form the largest component, with $291,637$ samples. In total, the dataset comprises $317.8$K samples, with $206.6$K for training, $63.6$K for validation, and $47.7$K for testing, providing a sufficient scale for learning spatial affordance.

\begin{figure}[h]
    \centering
    \vspace{-4pt}
    \includegraphics[width=0.50\linewidth]{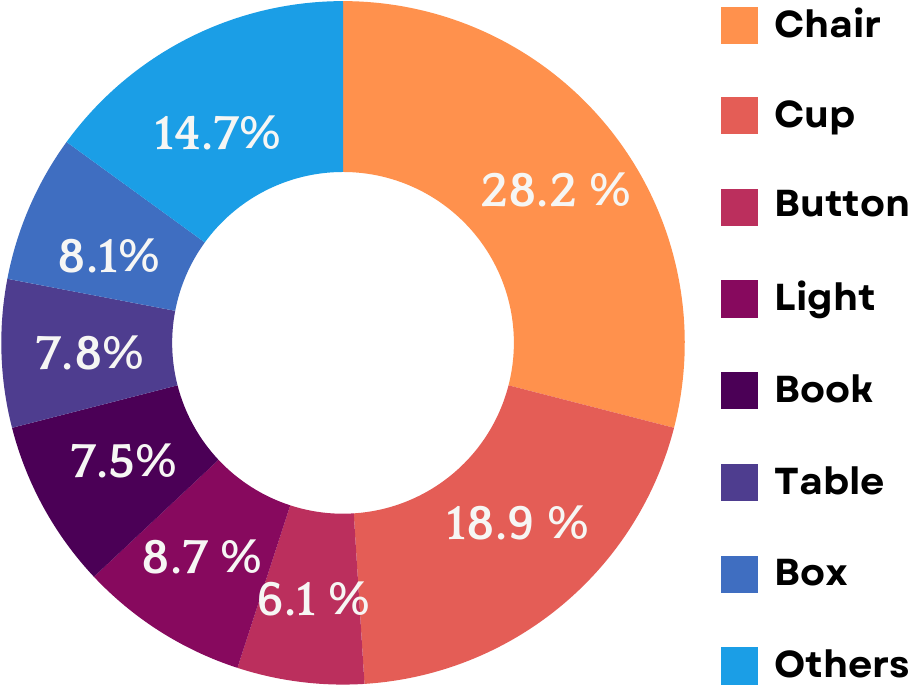}
    \vspace{-2pt}
    \caption{\textbf{Dataset statistics:} Statistics of objects in human-object interactions yielding affordances in our \textit{AffordBridge} dataset.}
    \label{fig:ObjDis}
    \vspace{-7pt}
\end{figure}

Fig.~\ref{fig:ObjDis} illustrates the object distribution across $689$ indoor 3D scenes. More specifically, chairs account for $28.2\%$, cups for $18.9\%$, and buttons for $8.7\%$, representing the most frequently interacted objects in the dataset. Lights contribute $7.5\%$, books $7.8\%$, tables $8.1\%$, and boxes $6.1\%$, each providing essential diversity for modeling functional affordances. The ``\textit{Others}'' category covers $14.7\%$ of the dataset. In Fig.~\ref{fig:ObjDis}, the object distribution demonstrates the diversity and balance of our \textit{AffordBridge} dataset, supporting both object-centric and scene-level affordance learning.

\vspace{-8pt}
\section{\textit{AffordMatcher}: Affordance Learning in 3D Scenes from Visual Signifiers}

\begin{figure*}[t]
    \centering
    \includegraphics[width=1.00\linewidth]{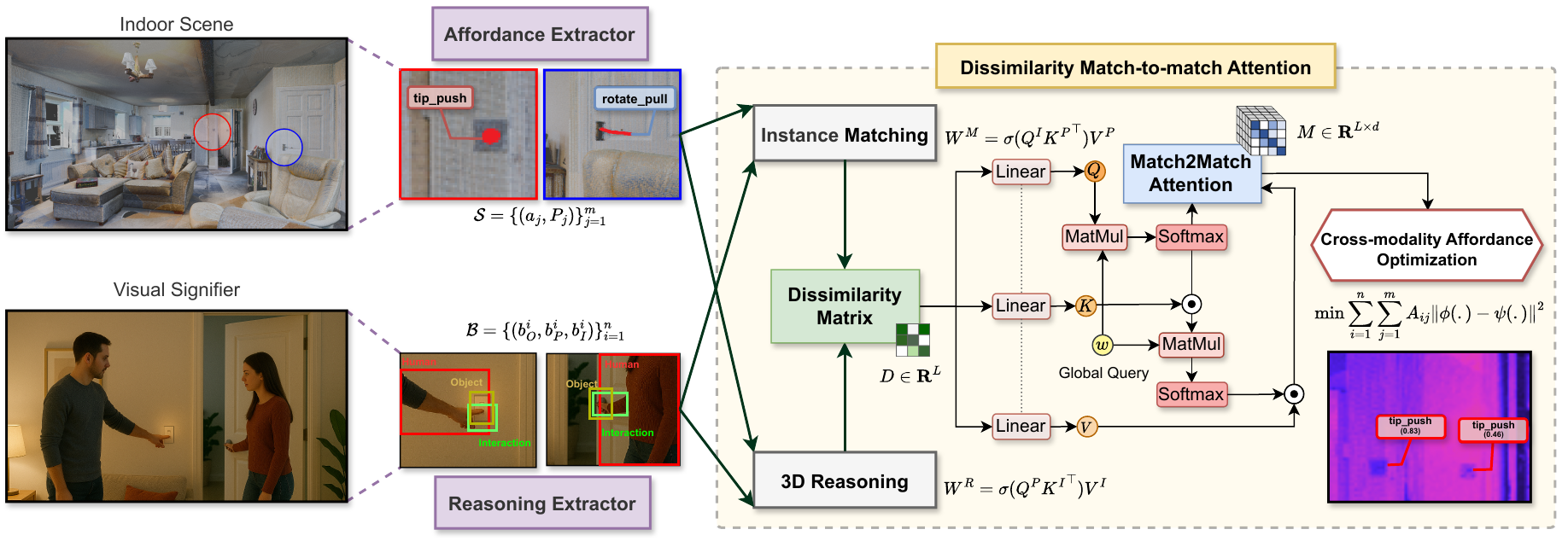}
    \vspace{-22pt}
    \caption{\textbf{Design architecture of \textit{AffordMatcher}:} Given a high-resolution voxelized scene point cloud and a visual signifier, \textit{AffordMatcher} reasons over these inputs for zero-shot affordance segmentation. The \textit{affordance extractor} identifies 3D interactable regions, while the \textit{reasoning extractor} encodes 2D human-object cues. Cross-modal alignment is achieved via instance matching through a dissimilarity matrix. The features from the dissimilarity matrix are thus optimized through match-to-match attention, followed by a zero-shot affordance optimization to localize actionable spatial regions that align with the given signifier.}
    \label{fig:baseline}
    \vspace{-14pt}
\end{figure*}

We formulate affordance grounding as an alignment problem between visual signifiers and 3D affordance regions. Our objective is to minimize the cross-modal feature discrepancy:
\begin{equation}
    \begin{aligned}
        \min_{\phi, \psi} \sum_{i=1}^{n} \sum_{j=1}^{m} A_{ij} \; \big| \phi(b_i) - \psi(a_j, P_j) \big|^{2}, \\
        \text{ s. t. } \sum_{j=1}^{m} A_{ij} = 1, \quad \forall i \in \{1, \dots, n\},
    \label{eq:matching_loss}
\end{aligned}
\end{equation}
where $\mathcal{P} = \{P_{j}\}_{j=1}^{m}$ denotes the point cloud composed of local regions $P_{j}$, and $I$ is the corresponding RGB image containing $n$ interaction cues $\{b_{i}\}_{i=1}^{n}$. In Eq.~\ref{eq:matching_loss}, the reasoning extractor $\phi: \mathbb{R}^{d_b} \rightarrow \mathbb{R}^d$ encodes each $b_{i}$ that represents human-part, object, and interaction features. Meanwhile, the affordance extractor $\psi: \mathbb{R}^{d_{a}} \times \mathbb{R}^{3} \rightarrow \mathbb{R}^{d}$ is mapping each 3D region $P_{j}$ and its action descriptor $a_{j}$ into the same embedding space. The weight $A_{ij}$ measures the confidence that $b_{i}$ aligns with the affordance instance $(a_{j}, P_{j})$.

\subsection{Instance Matching \& 3D Reasoning}
\label{subsec:matching}
Let $F_{P} \in \mathbb{R}^{m \times N_{P}}$ and $F_{I} \in \mathbb{R}^{n \times N_{I}}$ denote the feature representations of 3D candidate regions $\{M_{j}\}_{j=1}^{m}$ and visual signifiers $\{b_{i}\}_{i=1}^{n}$, where $N_{P}$ and $N_{I}$ represent the number of point features and the dimensionality of each visual signifier, respectively. We first project them into a shared space of dimension $N_{D}$ to obtain queries, keys, and values as: 
\begin{equation}
    \begin{aligned}
        Q^{(I)} &= F_I W_q^{(I)},\text{ }K^{(P)} = F_P W_k^{(P)},\text{ }V^{(P)} = F_P W_v^{(P)}, \\
        Q^{(P)} &= F_P W_q^{(P)},\text{ }K^{(I)} = F_I W_k^{(I)},\text{ }V^{(I)} = F_I W_v^{(I)}.
    \end{aligned}
    \label{eq:attention_comps}
\end{equation}
Based on Eq.~\ref{eq:attention_comps}, cross-modal attention is then applied bidirectionally to align 2D and 3D representations as follows:
\vspace{-7pt}
\begin{subequations}
    \vspace{-5pt}
    \begin{align}
        W^{(M)} &= \texttt{softmax}\left(Q^{(I)} {K^{(P)}}^{\top}\right) V^{(P)}, \\
        W^{(R)} &= \texttt{softmax}\left(Q^{(P)} {K^{(I)}}^{\top}\right) V^{(I)},
    \end{align}
    \label{eq:cross_modal_attention}
\end{subequations}
where $W^{(M)} \in \mathbb{R}^{n \times N_{D}}$ localizes spatial keypoint features guided by visual signifiers, and $W^{(R)} \in \mathbb{R}^{m \times N_D}$ captures reasoning feedback propagated from the 3D context.


\subsection{Dissimilarity Quantification}
From Eq.~\ref{eq:cross_modal_attention}, we compute the dissimilarity matrix $D \in \mathbb{R}^{n \times m}$ to quantify the cross-modal correspondence between visual and spatial features. Each entry $D_{ij} \in [0, 1]$ measures the cosine dissimilarity between the $i$-th and the $j$-th spatial features:
\begin{equation}
    D_{ij} = 1 - \max \left\{0, \, \frac{W^{(M)}_{i} \cdot W^{(R)}_{j}}{\|W^{(M)}_{i}\|_2 \, \|W^{(R)}_{j}\|_2}\right\},
    \label{eq:dissimilarity_matrix}
\end{equation}
where $\cdot$ is the inner product and $\|\cdot\|_{2}$ is the Euclidean norm.

The dissimilarity matrix $D$, with each entry as in Eq.~\ref{eq:dissimilarity_matrix}, is flattened into a length $L = nm$ and projected into an embedding of dimension $N_{X}$, yielding $X = DW_{X} \in \mathbb{R}^{L \times N_{X}}$. We then apply additive FastFormer-style self-attention~\citep{wu2021fastformer} over $(Q, K, V)$, defined to be $(XW_{q}, XW_{k}, XW_{v})$, as:
\begin{equation}
    Z = K \odot \left( \sigma(Q w_{q})^{\top} Q \right), \; M = V \odot \left( \sigma(Z w_{k})^{\top} Z \right),
    \label{eq:FastFormerM}
\end{equation}
where $\sigma(\cdot)$ denotes the element-wise sigmoid, $\odot$ represents the Hadamard product, and $w_{q}, w_{k} \in \mathbb{R}^{N_{X}}$ are learnable parameter vectors. Thus, a multi-head projection produces the final match matrix. With $M$ from Eq.~\ref{eq:FastFormerM}, we obtain:
\begin{equation}
    \mathcal{M} = \texttt{MultiHead}(M) \; W_{h} + b_{h},
    \label{eq:multihead_projection}
\end{equation}
where $\mathcal{M}$ is the \texttt{Match2Match} attention map, which is processed by the bounding-box and mask prediction heads. To address one-to-many correspondences as described, we apply a soft-thresholding mask on $D$. High-similarity pairs, $D_{ij} < 0.2$, are allowed to propagate multiple times through $\mathcal{M}$ to seek robust and stable matches in cluttered scenes.

\subsection{Cross-modality Affordance Learning}
To enable cross-modality affordance learning that solves Eq.~\ref{eq:matching_loss}, we align (\textit{i}) embedding normalization for global consistency (Eq.~\ref{eq:embed_loss}), followed by (\textit{ii}) semantic and geometric embeddings (Eq.~\ref{eq:align_loss}), (\textit{iii}) bidirectional mapping between modalities (Eq.~\ref{eq:bidirection_loss}), and (\textit{iv}) cross-modal attention dissimilarity (Eq.~\ref{eq:dissim_loss}). 

First, let $\phi_{i}: b_{i} \mapsto \mathbb{R}^{d}$ and $\psi_{j}: (a_{j}, P_{j}) \mapsto \mathbb{R}^{d}$ denote the projection heads for visual signifiers and spatial regions, respectively. Both embeddings are constrained to lie on the unit hypersphere, $\|\phi(b_{i})\|_{2} = 1$ and $\|\psi(a_{j}, P_{j})\|_{2} = 1$, to maintain feature and geometric consistency. The weighted sum of a normalization term and a regularization term results in the embedding regularization loss $\mathcal{L}_{\text{embed}}$:
\begin{equation}
    \alpha \left[ \sum_{i=1}^{n}(\|\phi_i\|_2 - 1)^2 + \sum_{j=1}^{m}(\|\psi_j\|_2 - 1)^2 \right] + \beta \sum_{\theta} \|\theta\|_F^2,
    \label{eq:embed_loss}
\end{equation}
where $\theta \in \Theta_\phi \cup \Theta_\psi$, $\Theta_{\phi}$ and $\Theta_{\psi}$ represent the trainable parameters of $\phi$ and $\psi$, respectively. Then, let $M_{ij}$ denote the FastFormer attention output (Eq.~\ref{eq:FastFormerM}) and $T_{ij}$ its pseudo-target from S-CLIP~\cite{mo2023s}, computed as a convex combination of CLIP text embeddings. The alignment loss $\mathcal{L}_{\mathrm{align}}$ is:
\begin{equation}
    \mathcal{L}_{\mathrm{align}} = \sum_{i=1}^{n}\sum_{j=1}^{m} A_{ij}\,\|M_{ij} - T_{ij}\|_2^2.
    \label{eq:align_loss}
\end{equation}
Next, with two linear projection heads, $g_{\text{ins}}, g_{\text{r}}: \mathbb{R}^d \rightarrow \mathbb{R}^d$, we further enforce bidirectional consistency $\mathcal{L}_{\text{bidir}}$:
\begin{equation}
    \mathcal{L}_{\text{bidir}} = \sum_{i,j} A_{ij} \left( \|g_{\text{ins}}(\phi_i) - \psi_j\|_2^2 + \|g_{\text{r}}(\psi_j) - \phi_i\|_2^2 \right)
    \label{eq:bidirection_loss}
\end{equation}
Lastly, we penalize the ReLU-clipped cosine dissimilarity between $W^{(M)}$ and $W^{(R)}$, yielding the dissimilarity loss that explicitly reduces the cross-modal attention $\mathcal{L}_{\text{dissim}}$:
\begin{equation}
    \mathcal{L}_{\text{dissim}} = \sum_{i,j} A_{ij} \left[ 1 - \frac{W^{(M)}_i \!\cdot\! W^{(R)}_j}{\|W^{(M)}_i\|_2 \|W^{(R)}_j\|_2} \right].
    \label{eq:dissim_loss}
\end{equation}

Overall, the training objective combines all losses with the empirically chosen set of weights $\{\alpha, \beta, \lambda, \gamma, \eta \}$:
\begin{equation}
    \mathcal{L}_{\text{total}} = \mathcal{L}_{\text{embed}} + \lambda\,\mathcal{L}_{\text{align}} + \gamma \mathcal{L}_{\text{bidir}} + \eta\,\mathcal{L}_{\text{dissim}}.
    \label{eq:total_loss}
\end{equation}
To train Eq.~\ref{eq:total_loss}, we employ the reasoning extractor $\phi$ and the affordance extractor $\psi$ with ViT-B/16~\cite{dosovitskiy2020imagevit16} and PointNet++~\cite{qi2017pointnet++}, respectively, each followed by two-layer MLP projection heads, with pose augmentation applied to $\phi$.

\section{Experiments \& Evaluations}

\subsection{Experiment Setup \& Baselines}
\textbf{Implementation Details.} In our experiments, the \textit{AffordMatcher} model is trained with RGB-point cloud and description text inputs for $100$ epochs on an NVIDIA RTX 3090 GPU with a batch size of $16$ and an initial learning rate of $10^{-4}$ decayed by $0.5$ every $30$ epochs. The RGB images are resized to $224 \times 224$ with visual augmentation, while 3D scenes are voxelized into a $64^{3}$ grid and segmented by a pre-trained 3D model to obtain binary-mask affordance candidates. Our evaluation follows the standardized zero-shot affordance segmentation metrics: mAP@$0.25$, mAP@$0.50$, and mAP averaged over IoU thresholds from $0.50$ to $0.95$.

\noindent \textbf{Baselines.} We benchmark \textit{AffordMatcher} against state-of-the-art baselines in functional adaptations of 3D instance segmentation methods, including Mask3D-F, SoftGroup-F, and OpenMask3D-F, as listed in~\cite{delitzas2024scenefun3d}, and full pipelines, including AffordPose-DGCNN~\cite{jian2023affordpose}, 3DAffordanceNet~\cite{deng20213d}, PIAD~\cite{yang2023grounding}, LASO~\cite{li2024laso}, and Ego-SAG~\cite{liu2024grounding}. 

\subsection{Quantitative Results}
Table~\ref{tab:3daffordance} reports the performance of our method against state-of-the-art methods on functionality affordance segmentation. Specifically, \textit{AffordMatcher} achieves an overall mAP of $53.4$, outperforming the second-best baseline by $7.8$ while exhibiting superior localization of functionally relevant regions among both low and high IoU thresholds. The improvement demonstrates the effectiveness of visually guided reasoning in improving spatial affordance localization in 3D scenes, given visual signifiers, under zero-shot settings.

\begin{table}[h]
    \centering
    \vspace{-4pt}
    \resizebox{\linewidth}{!}{
    \begin{tabular}{r|ccc|cc}
        \toprule
        \diagbox{Method}{Metric} & \makecell{mAP} & \makecell{mAP\\@0.25} & \makecell{mAP\\@0.50} & \makecell{No.\\Params} & \makecell{Inference Speed \\ (ms / sample)} \\
        \midrule \midrule
        Mask3D-F~\cite{delitzas2024scenefun3d} & 41.2 & 58.6 & 47.1 & 19.0M & \underline{126.2} \\ 
        SoftGroup-F~\cite{delitzas2024scenefun3d} & 43.9 & 60.8 & 49.3 & 30.4M & 288.0 \\ 
        OpenMask3D-F~\cite{delitzas2024scenefun3d} & \underline{45.6} & \underline{62.1} & \underline{51.0} & 39.7M & 315.1 \\
        \midrule
        APose-DGCNN~\cite{jian2023affordpose} & 29.7 & 47.6 & 34.8 & \textbf{12.5M} & 140.2 \\ 
        3DAffordanceNet~\cite{AffordanceNet18} & 34.2 & 51.3 & 39.6 & \underline{15.0M} & 180.4 \\ 
        PIAD~\cite{yang2023grounding} & 26.1 & 44.7 & 30.5 & 23.0M & 160.9 \\ 
        LASO~\cite{li2024laso} & 37.5 & 54.2 & 42.6 & 21.4M & 130.4 \\ 
        Ego-SAG~\cite{liu2024grounding} & 40.3 & 56.7 & 45.1 & 24.8M & 175.3 \\ 
        \midrule
        \rowcolor{gray!25} \textbf{\textit{AffordMatcher} (Ours)} & \textbf{53.4} & \textbf{69.7} & \textbf{59.5} & 20.7M & \textbf{112.5} \\
        \bottomrule
    \end{tabular}
    }
    \vspace{-5pt}
    \caption{\textbf{Quantitative results:} Performance comparisons of \textit{AffordMatcher} and state-of-the-art methods~\cite{delitzas2024scenefun3d, jian2023affordpose, AffordanceNet18, yang2023grounding, li2024laso, liu2024grounding} in terms of mAP, mAP@$0.25$, mAP@$0.50$, number of parameters (in millions), and inference speed (in milliseconds per sample).}
    \vspace{-5pt}
    \label{tab:3daffordance}
\end{table}

Also shown in Table~\ref{tab:3daffordance}, \textit{AffordMatcher} attains high accuracy while maintaining computational efficiency. The model contains $20.7$ million parameters; for example, fewer than OpenMask3D-F, while achieving faster inference at $112.5$ milliseconds per sample. The balanced trade-off between accuracy and efficiency illustrates the scalability of \textit{AffordMatcher} for large-scale and near real-time 3D scene affordance localization.

\begin{figure*}[t]
    \centering
    \setlength\tabcolsep{2pt} 
    \renewcommand\arraystretch{0.6} 
    \begin{tabular}{cccc}
        \begin{tabular}{@{}c@{}}
            \includegraphics[width=0.23\linewidth]{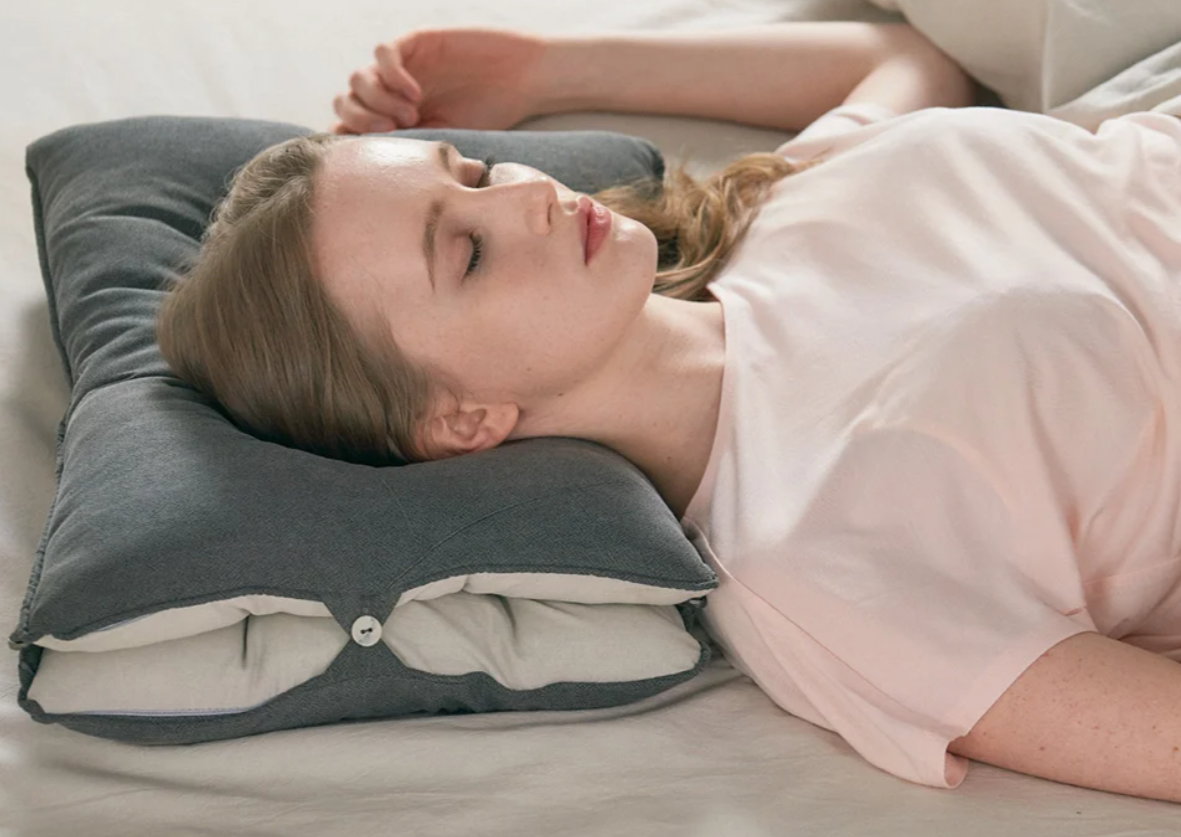} \\[0.1cm]
            \small{\textbf{(a)} Visual signifer}
        \end{tabular}
        &
        \begin{tabular}{@{}c@{}}
            \includegraphics[width=0.23\linewidth]{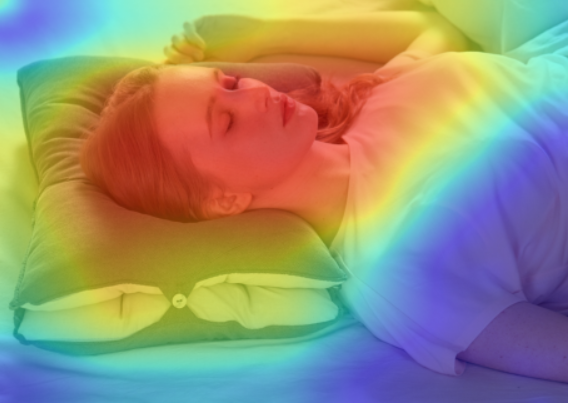} \\[0.1cm]
            \small{\textbf{(b)} Attention over visual signifer}
        \end{tabular}
        &
        \begin{tabular}{@{}c@{}}
            \includegraphics[width=0.23\linewidth]{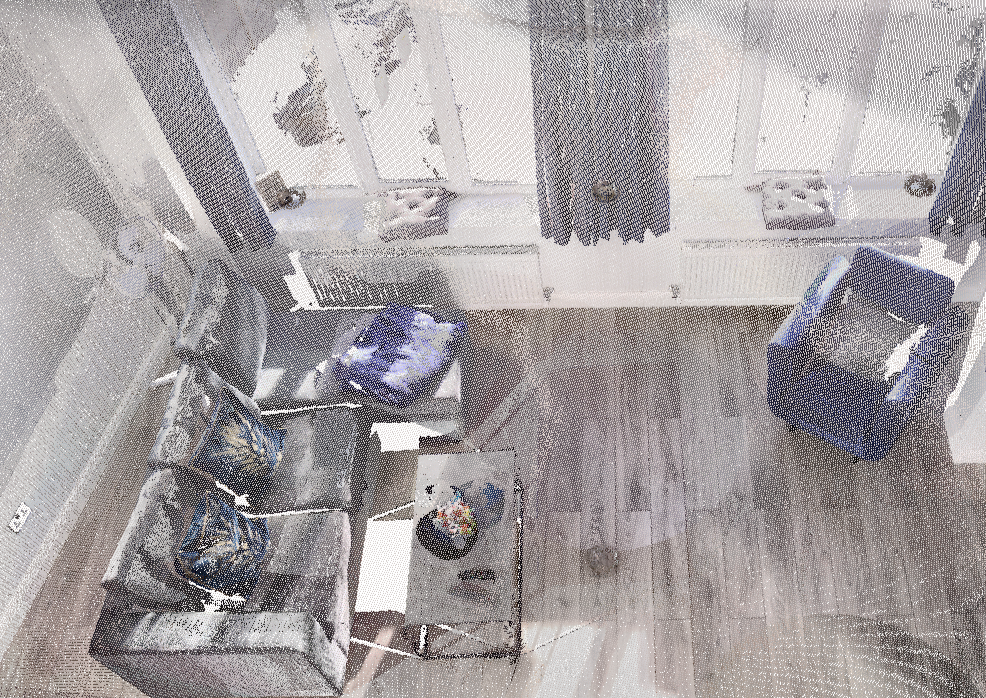} \\[0.1cm]
            \small{\textbf{(c)} High-resolution voxelized scene}
        \end{tabular}
        &
        \begin{tabular}{@{}c@{}}
            \includegraphics[width=0.23\linewidth]{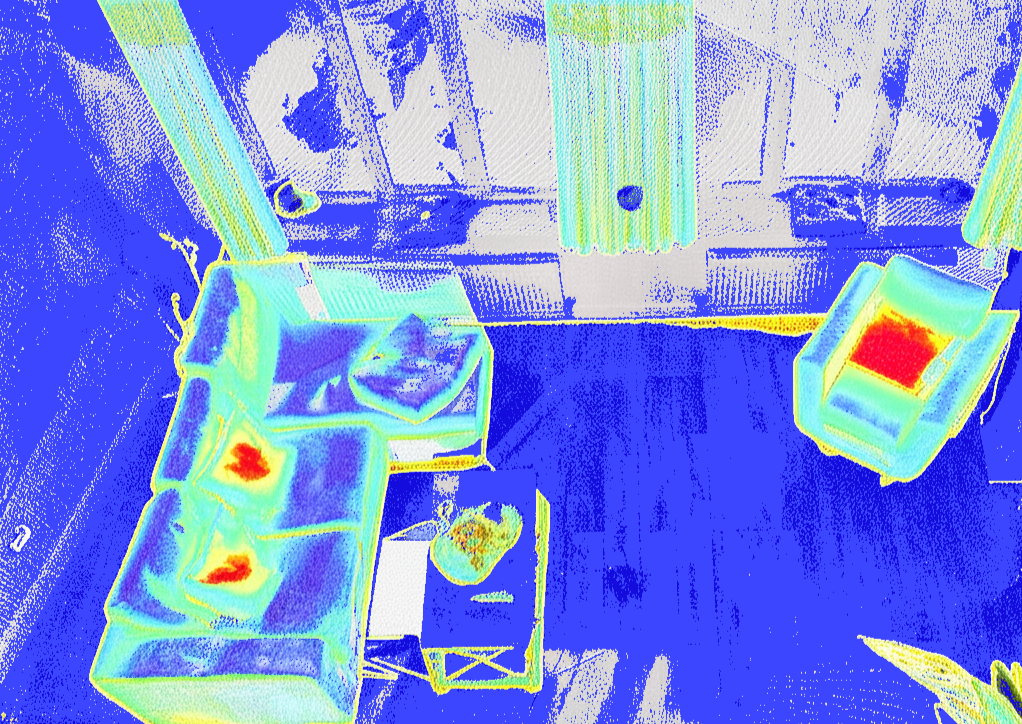} \\[0.1cm]
            \small{\textbf{(d)} Attention over the given scene}
        \end{tabular}
    \end{tabular}
    \vspace{-7pt}
    \caption{\textbf{Attention visualization:} From the visual signifier in the RGB image and the text ``\textit{Rest on Pillow}'', \textit{AffordMatcher} focuses on the pillow area in the RGB image and correctly localizes the corresponding affordance regions in the high-resolution voxelized indoor scene.}
    \vspace{-18pt}
    \label{fig:attention}
\end{figure*}

\begin{table}[t]
    \centering
    \vspace{4pt}
    \resizebox{0.90\linewidth}{!}{
    \begin{tabular}{r|ccc}
        \toprule
        \diagbox{Variant of Inputs}{Metric} & \makecell{mAP} & \makecell{mAP\\@0.25} & \makecell{mAP\\@0.50} \\
        \midrule \midrule
        pruning RGB image inputs & 37.3 & 52.7 & 42.1 \\
        inpainting human-object interactions & 40.9 & 56.2 & 45.3 \\
        without point cloud downsampling & \underline{48.7} & \underline{65.1} & \underline{54.2} \\
        fine-tuning with PIAD objects & 45.3 & 61.8 & 50.6 \\
        \midrule
        \rowcolor{gray!25} \textbf{\textit{AffordMatcher} (Ours)} & \textbf{53.4} & \textbf{69.7} & \textbf{59.5} \\
        \bottomrule
    \end{tabular}
    }
    \vspace{-4pt}
    \caption{\textbf{Ablation on different input modalities:} Removing visual inputs severely degrades performance, followed by inpainting human-object interactions, fine-tuning with PIAD objects, and thus using raw point cloud. The full \textit{AffordBridge} achieves the highest accuracy through integrated 2D-3D reasoning.}
    \vspace{-12pt}
    \label{tab:ablationInteraction}
\end{table}

\subsection{Ablation Study}
\textbf{Input Guidance Analysis.} As shown in Table~\ref{tab:ablationInteraction}, we conduct experiments to assess whether the observed performance gains originate from interaction reasoning rather than object recognition. We found that eliminating the 2D branch leads to a significant decrease in mAP to $37.3$, indicating the critical role of visual cues. Meanwhile, removing humans and hands from RGB images via inpainting~\citep{telea2004image} lowers the mAP to $40.9$, confirming that action semantics contribute to interaction reasoning. Using raw point clouds of more than $500,000$ points reduces the mAP to $48.7$ due to memory constraints. Fine-tuning on the object-centric PIAD dataset produces $45.3$ of mAP, validating the advantage of modeling scene-level affordance cues over isolated object interactions.


\begin{table}[h]
    \centering 
    \vspace{-8pt}
    \resizebox{0.90\linewidth}{!}{
    \begin{tabular}{cccc|ccc}
        \toprule
        $\mathcal{L}_{\text{align}}$ & $\mathcal{L}_{\text{dissim}}$ & $\mathcal{L}_{\text{embed}}$ & $\mathcal{L}_{\text{bidir}}$ & mAP & \makecell{mAP\\@0.25} & \makecell{mAP\\@0.50} \\ 
        \midrule \midrule
        & & & & 37.3 & 52.7 & 42.1 \\ 
        \midrule
        \checkmark & & & & 40.9 & 56.2 & 45.3 \\
        \checkmark & \checkmark & & & 44.1 & 60.0 & 48.7 \\ 
        \checkmark & \checkmark & \checkmark & & \underline{47.8} & \underline{63.5} & \underline{53.0} \\ 
        \rowcolor{gray!25} \checkmark & \checkmark & \checkmark & \checkmark & \textbf{53.4} & \textbf{69.7} & \textbf{59.5} \\ 
        \bottomrule
    \end{tabular}
    }
    \vspace{-4pt}
    \caption{\textbf{Ablation on loss components:} Gradual inclusion of each loss enhances performance, with the full objective achieving the best results. The first row indicates the baseline, which corresponds to the semantic affordance objective~\cite{delitzas2024scenefun3d}.}
    \vspace{-5pt}
    \label{tab:ablationLoss}
\end{table}

\noindent \textbf{Loss Component Analysis.} Table~\ref{tab:ablationLoss} reports the contribution of each loss component. Beginning with a standard semantic baseline, the inclusion of $\mathcal{L}_{\text{align}}$ and $\mathcal{L}_{\text{dissim}}$ yields significant improvements, showcasing the importance of cross-instance correspondence modeling. Adding the bidirectional and regularization losses further enhances performance, resulting in a cumulative mAP gain of $16.1$, which shows that the complete objective effectively learns structured and discriminative affordance representations.

\begin{figure}[t]
    \begin{minipage}[ht]{0.98\linewidth}
        \centering
        \vspace{4pt}
        \setlength{\tabcolsep}{1em} 
        {\renewcommand{\arraystretch}{1}
            \begin{tabular}{c c}
            \includegraphics[width=0.40\linewidth]{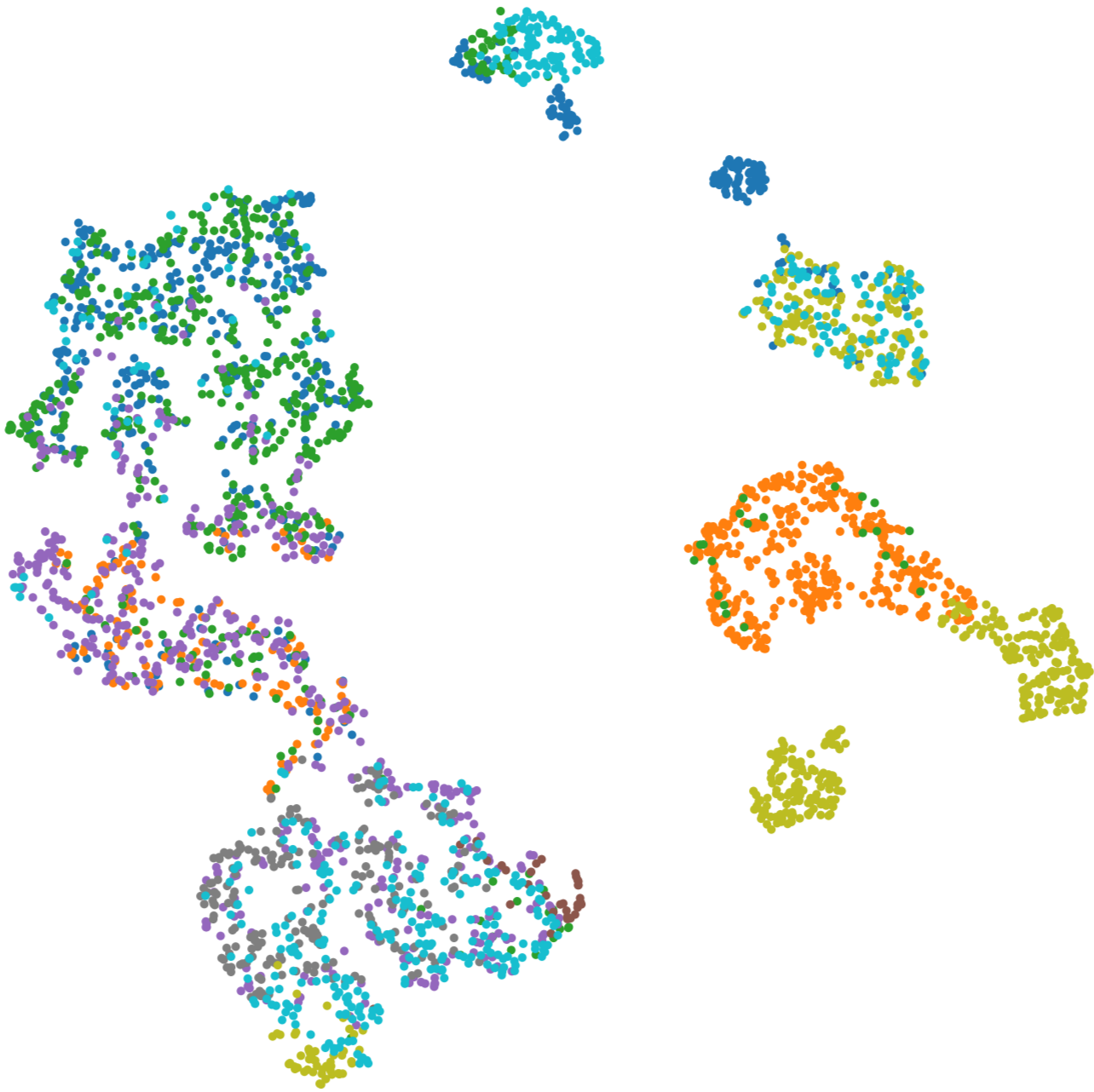} &
            \includegraphics[width=0.40\linewidth]{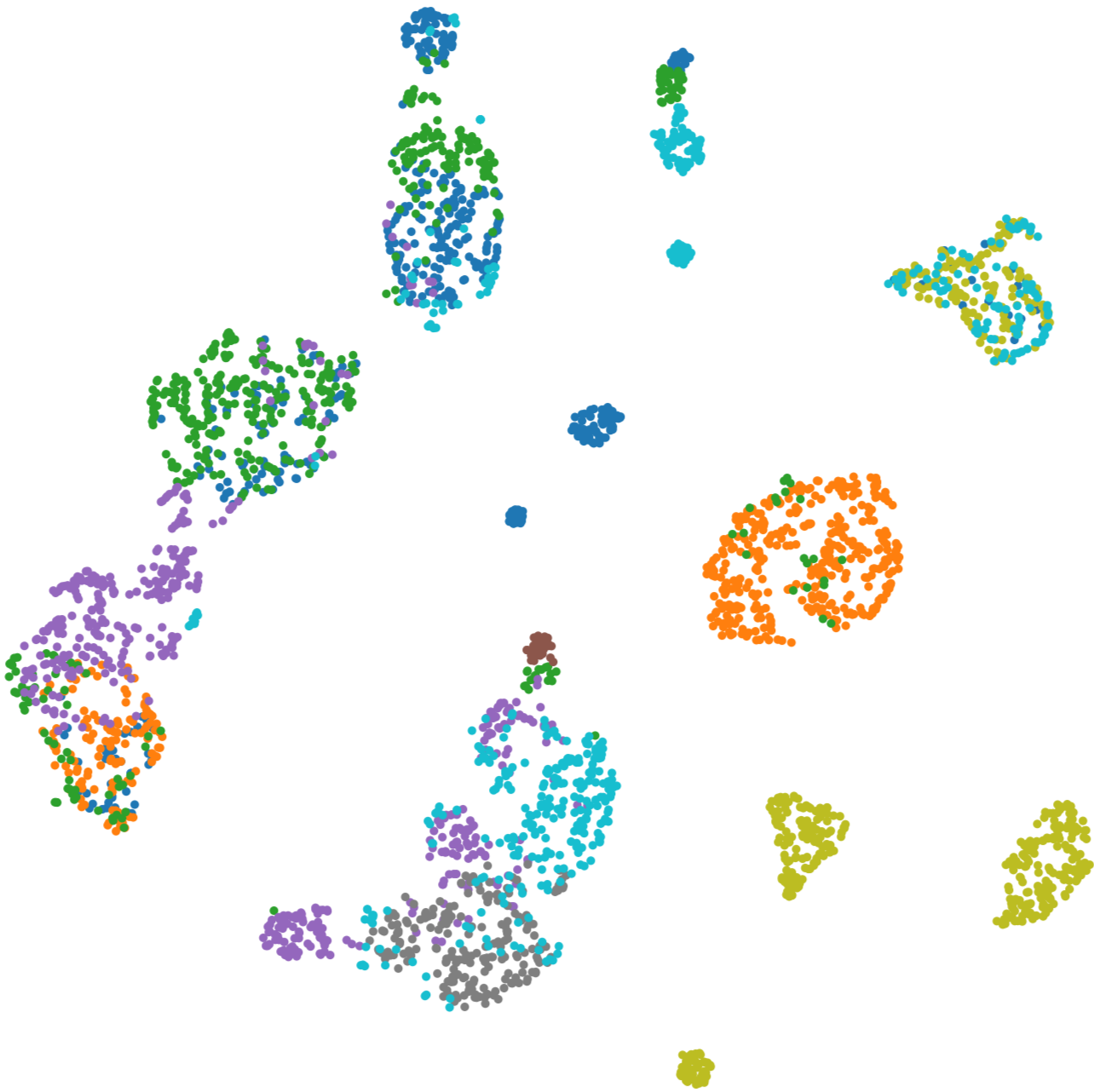} \\
            \begin{tabular}{@{}c@{}}\small{\textbf{(a)} without visual reasoning}\end{tabular} &
            \begin{tabular}{@{}c@{}}\small{\textbf{(b)} with visual reasoning}\end{tabular} 
            \end{tabular}
        }
        \vspace{-7pt}
        \caption{\textbf{t-SNE visualization:} Visual reasoning produces more compact and well-separated clusters among different affordance types compared to without visual reasoning.}
        \label{fig:tSNE}
    \end{minipage}
    \vspace{-18pt}
\end{figure}

\begin{figure}[b]
    \centering
    \vspace{-15pt}
    \includegraphics[width=1.00\linewidth]{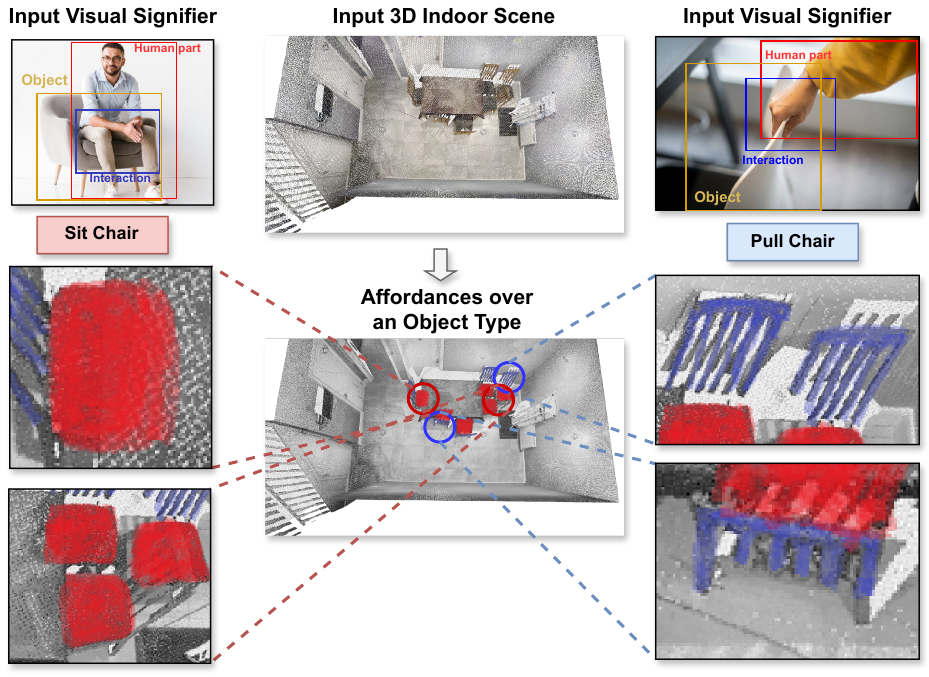}
    \vspace{-18pt}
    \caption{\textbf{Visualization of distinct affordances:} ``\textit{Sit}'' and ``\textit{Pull}'' cues on the same chair activate different 3D regions, showcasing that \textit{AffordMatcher} adapts attention to interaction semantics.}
    \label{fig:AffPredwDiffAction}
\end{figure}

\noindent \textbf{Attention Visualization.} Fig.~\ref{fig:attention} illustrates how the learned attention maps transfer reasoning cues across modalities. For the ``\textit{Rest on Pillow}'' example, the visual signifier concentrates on the pillow region. Through this, its spatial attention is able to emphasize the corresponding voxels on seating surfaces, confirming consistent cross-modal learning for human-object affordance localization.

\begin{figure*}[!ht]
    \centering
    \resizebox{0.90\linewidth}{!}{
    \setlength\tabcolsep{2pt} 
    \renewcommand\arraystretch{0.8} 
    \begin{tabular}{cccc}
        \begin{tabular}{@{}c@{}}
            \includegraphics[width=0.12\linewidth]{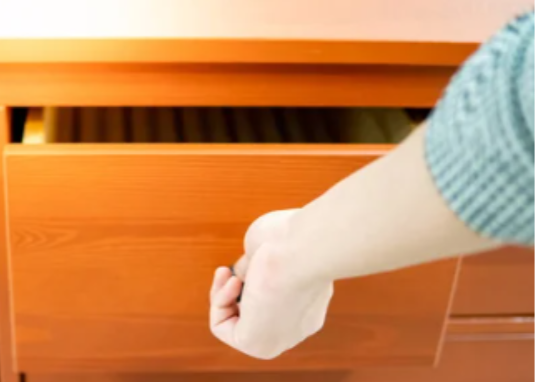} \\
            \includegraphics[width=0.12\linewidth]{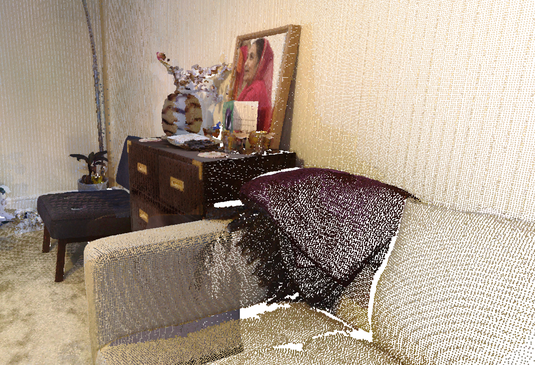}
        \end{tabular}
        &
        \begin{tabular}{@{}c@{}}
            $\langle$\textcolor{blue}{\texttt{Open}}$\rangle\ \langle$--$\rangle$ \\ 
            \includegraphics[width=0.22\linewidth]{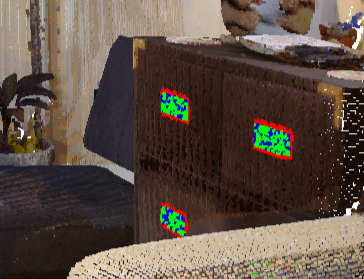}
        \end{tabular}
        &
        \begin{tabular}{@{}c@{}}
            $\langle$\textcolor{blue}{\texttt{Open}}$\rangle\ \langle$\textcolor{red}{\texttt{Case}}$\rangle$ \\ 
            \includegraphics[width=0.22\linewidth]{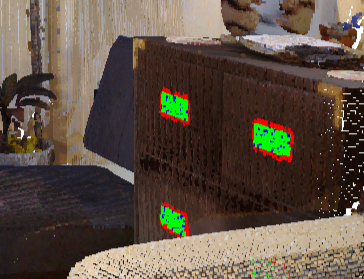}
        \end{tabular}
        &
        \begin{tabular}{@{}c@{}}
            $\langle$\textcolor{blue}{\texttt{Open}}$\rangle\ \langle$\textcolor{blue}{\texttt{Drawer}}$\rangle$ \\ 
            \includegraphics[width=0.22\linewidth]{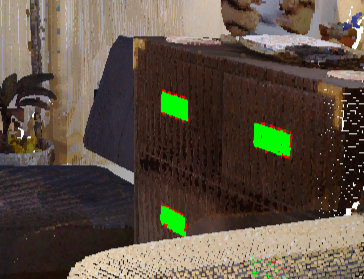}
        \end{tabular}
        \\[0.2cm] 

        \begin{tabular}{@{}c@{}}
            \includegraphics[width=0.12\linewidth]{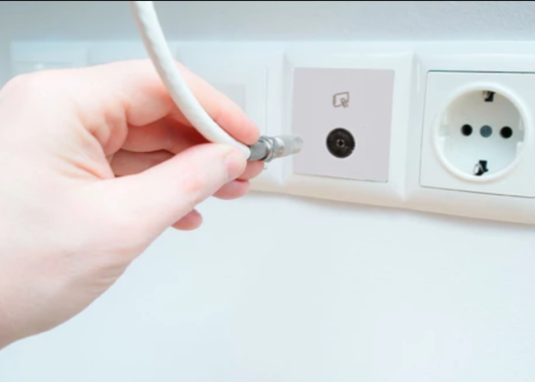} \\
            \includegraphics[width=0.12\linewidth]{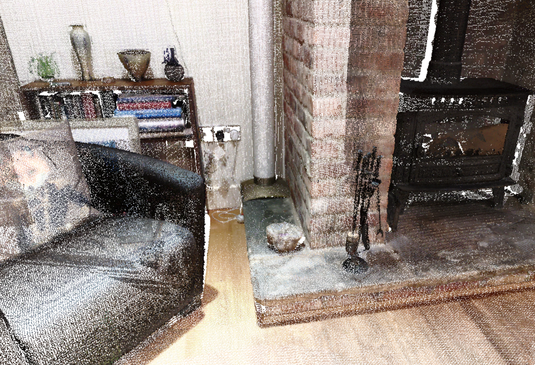}
        \end{tabular}
        &
        \begin{tabular}{@{}c@{}}
            $\langle$\textcolor{blue}{\texttt{Plug}}$\rangle\ \langle$--$\rangle$ \\ 
            \includegraphics[width=0.22\linewidth]{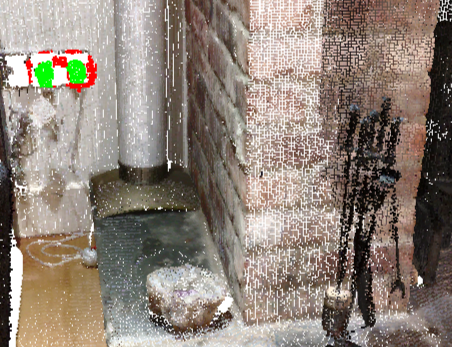}
        \end{tabular}
        &
        \begin{tabular}{@{}c@{}}
            $\langle$\textcolor{red}{\texttt{Put}}$\rangle\ \langle$\textcolor{blue}{\texttt{Jack}}$\rangle$ \\ 
            \includegraphics[width=0.22\linewidth]{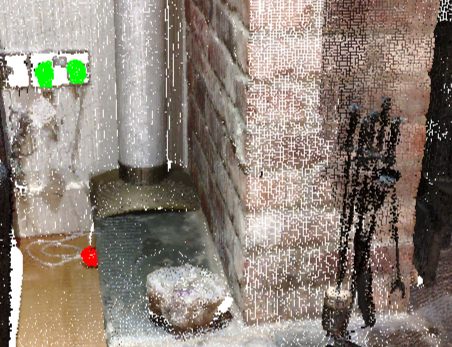}
        \end{tabular}
        &
        \begin{tabular}{@{}c@{}}
            $\langle$\textcolor{blue}{\texttt{Plug}}$\rangle\ \langle$\textcolor{blue}{\texttt{Jack}}$\rangle$ \\ 
            \includegraphics[width=0.22\linewidth]{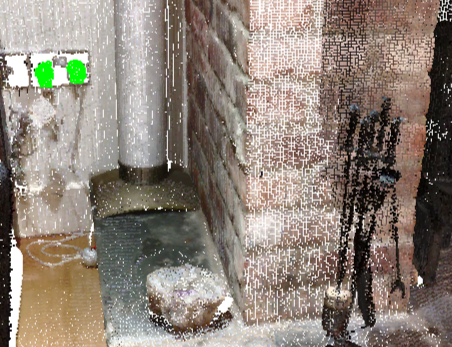} 
        \end{tabular}
        \\[0.2cm] 

        \begin{tabular}{@{}c@{}}
            \includegraphics[width=0.12\linewidth]{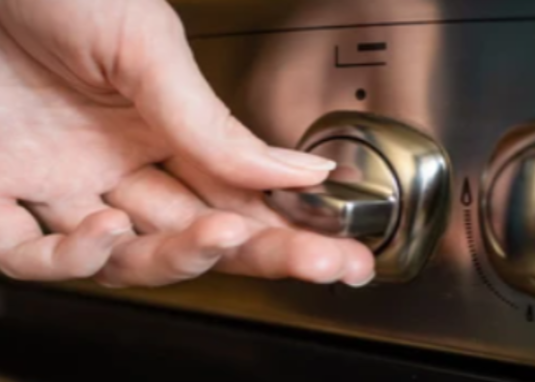} \\
            \includegraphics[width=0.12\linewidth]{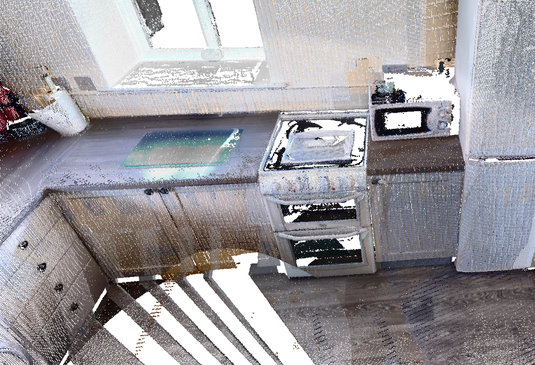}
        \end{tabular}
        &
        \begin{tabular}{@{}c@{}}
            $\langle$\textcolor{blue}{\texttt{Rotate}}$\rangle\ \langle$--$\rangle$ \\ 
            \includegraphics[width=0.22\linewidth]{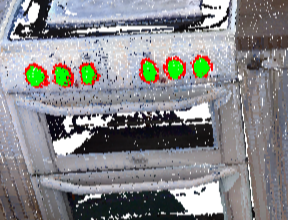} \\ 
            \small{PIAD~\cite{yang2023grounding}}
        \end{tabular}
        &
        \begin{tabular}{@{}c@{}}
            $\langle$\textcolor{blue}{\texttt{Rotate}}$\rangle\ \langle$\textcolor{blue}{\texttt{Button}}$\rangle$ \\ 
            \includegraphics[width=0.22\linewidth]{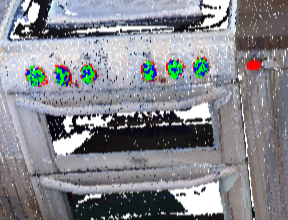} \\ 
            \small{Ego-SAG~\cite{liu2024grounding}}
        \end{tabular}
        &
        \begin{tabular}{@{}c@{}}
            $\langle$\textcolor{blue}{\texttt{Rotate}}$\rangle\ \langle$\textcolor{blue}{\texttt{Button}}$\rangle$ \\ 
            \includegraphics[width=0.22\linewidth]{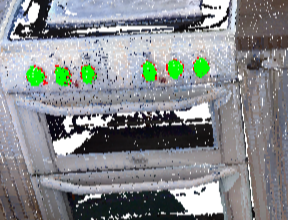} \\ 
            \small{\textbf{\textit{AffordMatcher} (Ours)}}
        \end{tabular}
    \end{tabular}
    }
    \vspace{-7pt}
    \caption{\textbf{Qualitative results:} Our affordance‐mask prediction, compared with other baselines. The first column illustrates the inputs, including visual signifiers and 3D scenes. The remaining columns show the affordance segmentation results, including extracted actions over different methods, where \textcolor{blue}{\textbf{blue texts}} indicate correct affordance actions and \textcolor{red}{\textbf{red texts}} represent wrong ones. In the point clouds, \textcolor{green}{\textbf{green areas}} denote correct affordance localization,  \textcolor{red}{\textbf{red}} and \textcolor{blue}{\textbf{blue areas}} denote false positives and false negatives, respectively.}
    \vspace{-16pt}
    \label{fig:3DAffVis}
\end{figure*}

\noindent \textbf{Reasoning Analysis.} The t-SNE embeddings~\cite{van2008visualizing} in Fig.~\ref{fig:tSNE} reveal that the reasoning module on Sec.~\ref{subsec:matching} produces compact and well-separated clusters, indicating improved affordance discriminability. Meanwhile, Fig.~\ref{fig:AffPredwDiffAction} further demonstrates reasoning adaptability by visualizing distinct interaction cues, such as ``\textit{Sit}'' and ``\textit{Pull}'', applied to the same object. For the ``\textit{Sit}'' action, attention focuses on the seat cushion and the frontal area of the chair; whereas for the ``\textit{Pull}'' action, the attention then shifts toward the upper back and armrest regions, demonstrating that \textit{AffordMatcher} can dynamically adjust its focus according to given visual signifiers while maintaining a consistent geometric understanding.

\noindent\textbf{Qualitative Results.} Fig.~\ref{fig:3DAffVis} presents qualitative comparisons between our \textit{AffordMatcher} and PIAD~\cite{yang2023grounding} together with Ego-SAG~\cite{liu2024grounding}. PIAD tends to under-segment affordance regions, often missing fine interaction details, while Ego-SAG often over-segments them, producing overly broad or redundant affordance masks that lack spatial precision. Notably, \textit{AffordMatcher} is able to generate compact and accurate affordance masks that suit both coarse and fine-grained interaction parts, such as knobs and prongs, while achieving higher spatial precision.

\noindent \textbf{Limitations.} Although \textit{AffordMatcher} demonstrates strong cross-modality affordance learning capability, it faces challenges with memory and scalability in highly detailed scenes, resulting in increased computational costs. Some errors also occur, such as overlapping affordances or unclear actions (please see details in our Supplementary Material), which reveal limits in disambiguation and spatial reasoning.

\vspace{-3pt}
\section{Conclusions}
In this work, we introduce \textit{AffordMatcher}, an affordance learning method for spatial affordance localization in high-resolution voxelized indoor scenes from visual signifiers through sophisticated cross-modal reasoning. By leveraging the \textit{AffordBridge} dataset and a dissimilarity-based match-to-match attention mechanism, \textit{AffordMatcher} achieves robust zero-shot affordance segmentation across diverse scenes. Experimental results demonstrate consistent gains over state-of-the-art methods in both accuracy and efficiency, validating the effectiveness of reasoning-guided affordance learning. We reserve the task of extending \textit{AffordMatcher} to temporal and interactive scenarios, enabling dynamic affordance reasoning in real-world robotic systems and embodiments.

{\small
    \bibliographystyle{ieeenat_fullname}
    \bibliography{egbib}
}

\clearpage
\setcounter{page}{1}
\twocolumn[{%
\renewcommand\twocolumn[1][]{#1}%
\maketitlesupplementary
\begin{center}
    \setlength{\tabcolsep}{2pt}
    \centering
    \begin{tabular}{cccc}
\shortstack{\includegraphics[width=0.32\linewidth, height=0.19\linewidth]{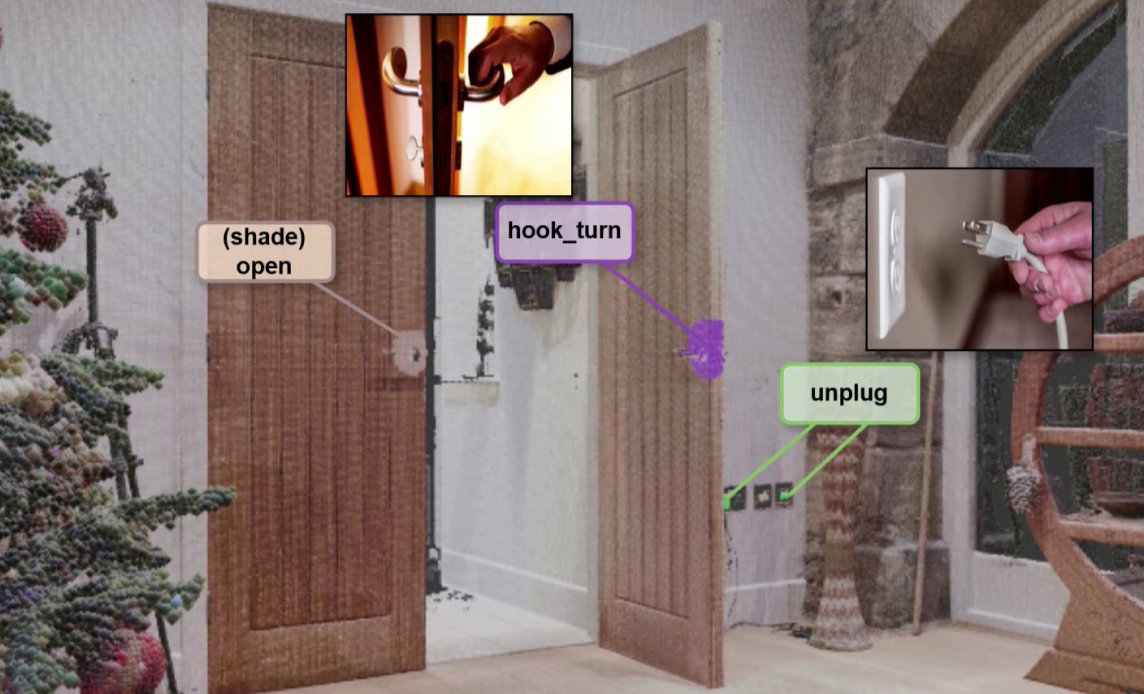}}&
\shortstack{\includegraphics[width=0.32\linewidth, height=0.19\linewidth]{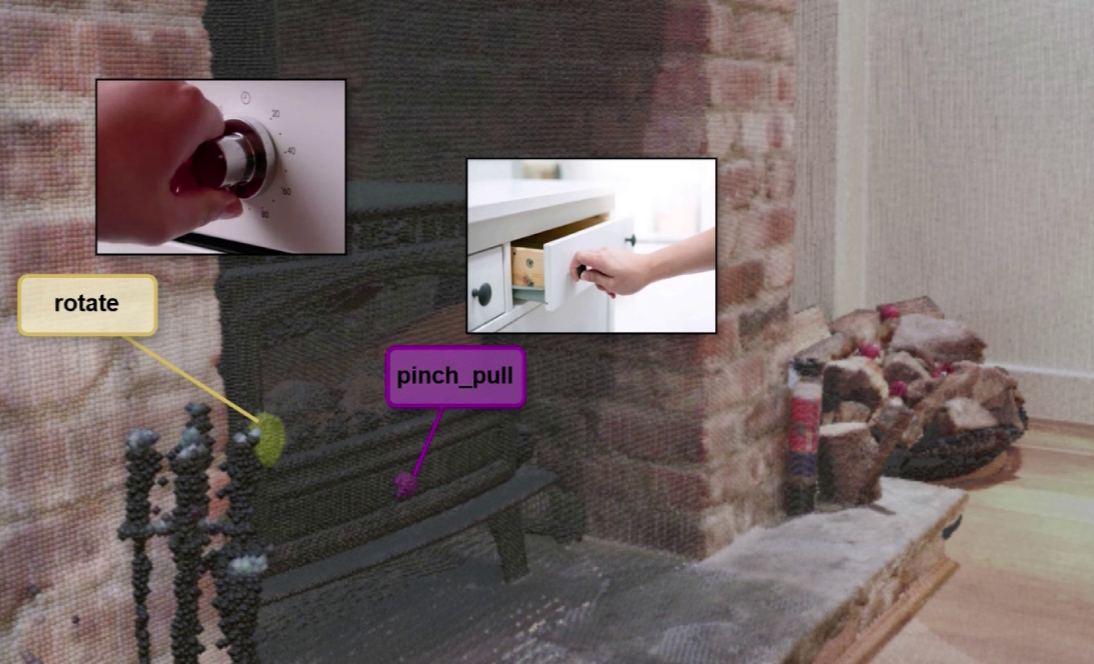}}&  \shortstack{\includegraphics[width=0.32\linewidth, height=0.19\linewidth]{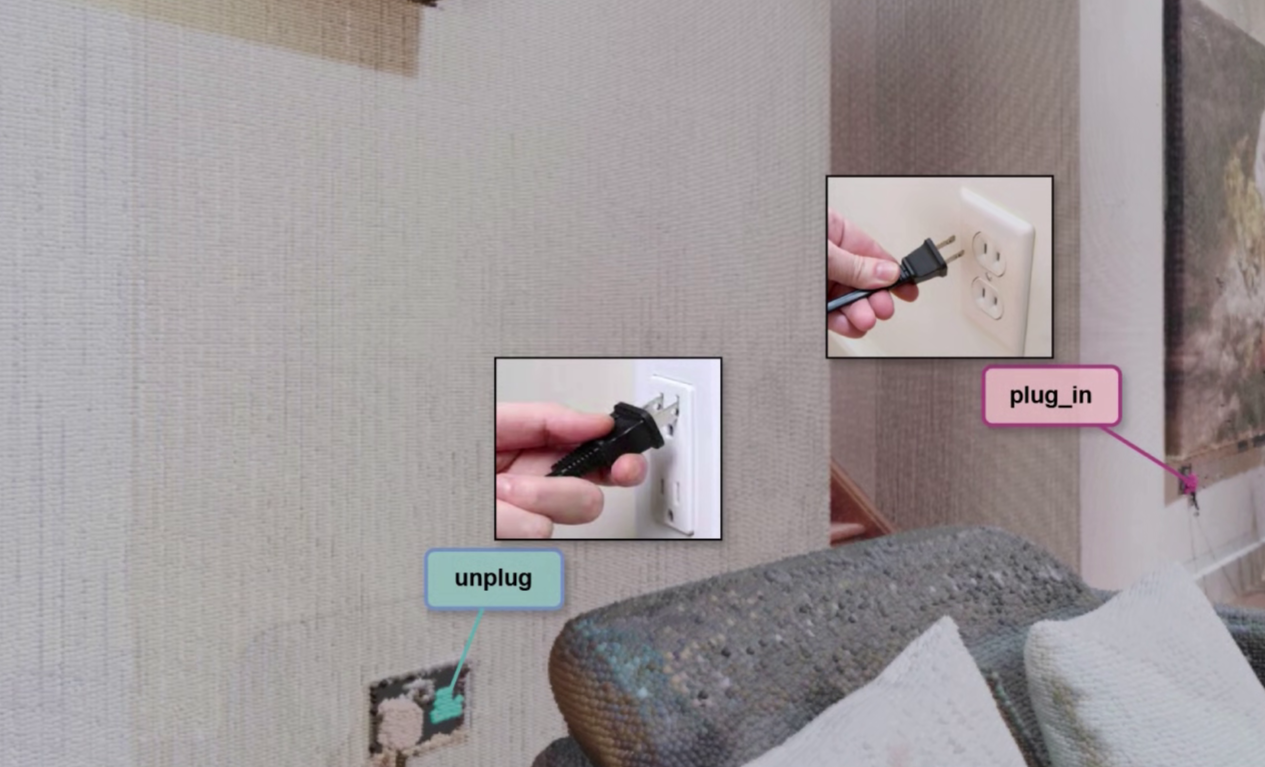}}\\[0pt]
\shortstack{\includegraphics[width=0.32\linewidth, height=0.19\linewidth]{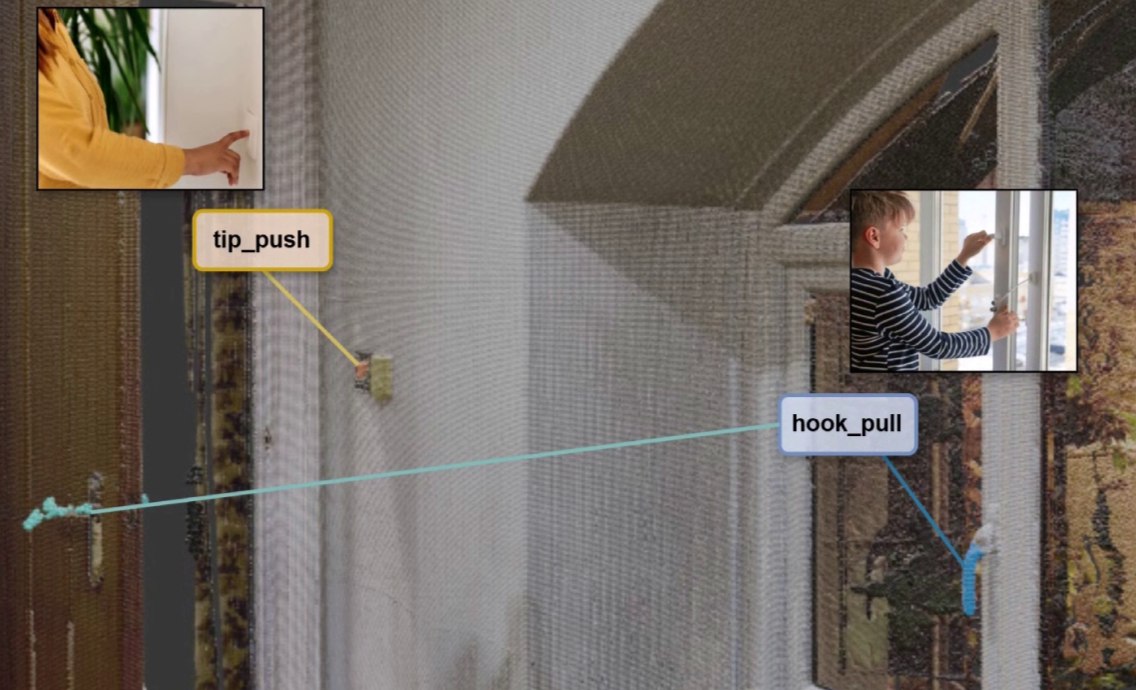}}&
\shortstack{\includegraphics[width=0.32\linewidth, height=0.19\linewidth]{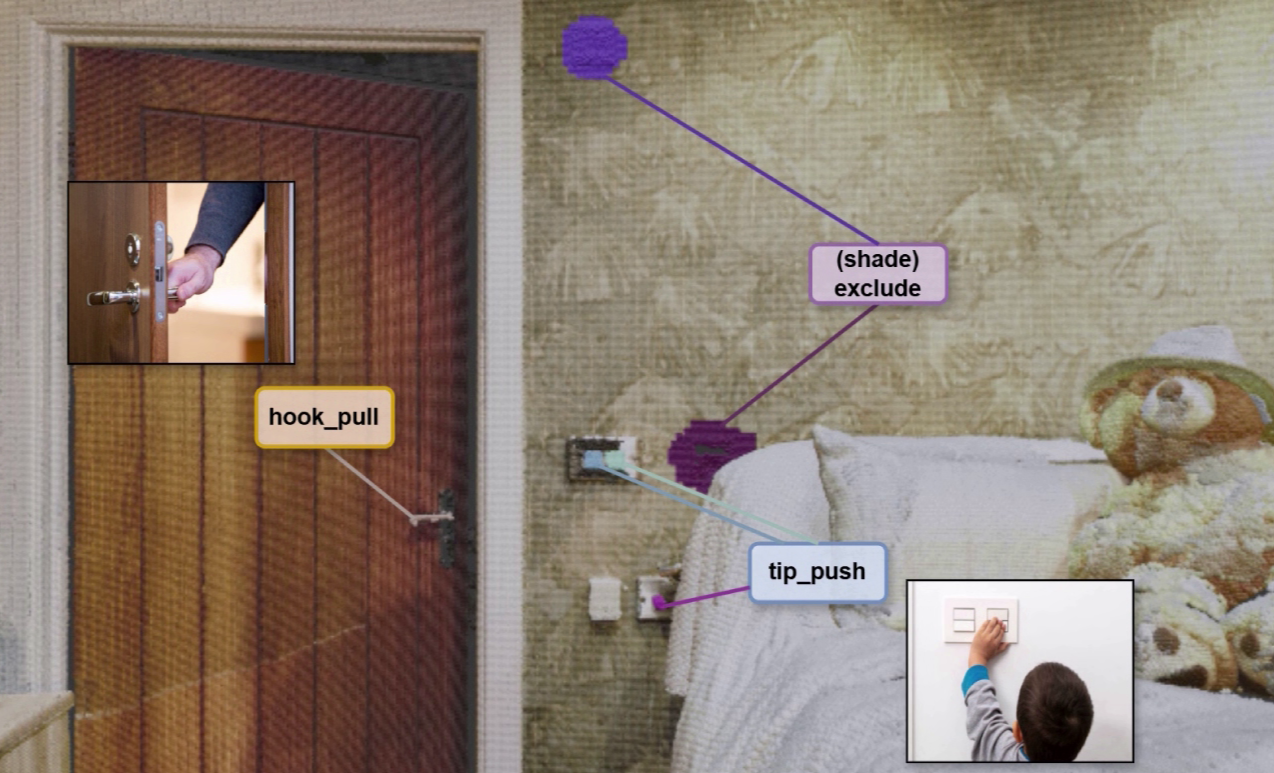}}&  \shortstack{\includegraphics[width=0.32\linewidth, height=0.19\linewidth]{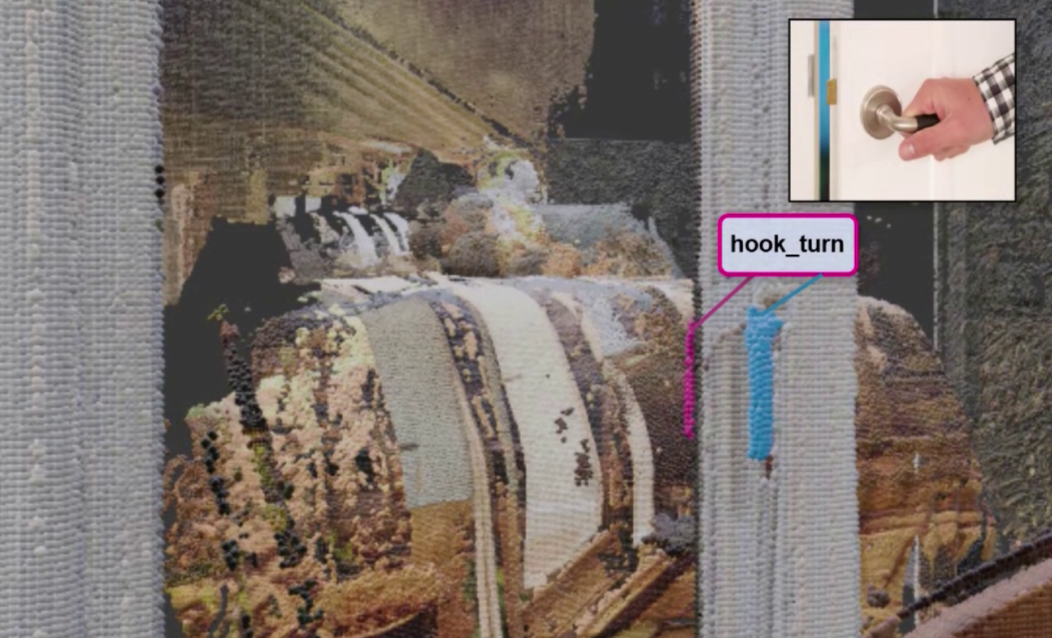}}\\[0pt]

\shortstack{\includegraphics[width=0.32\linewidth, height=0.19\linewidth]{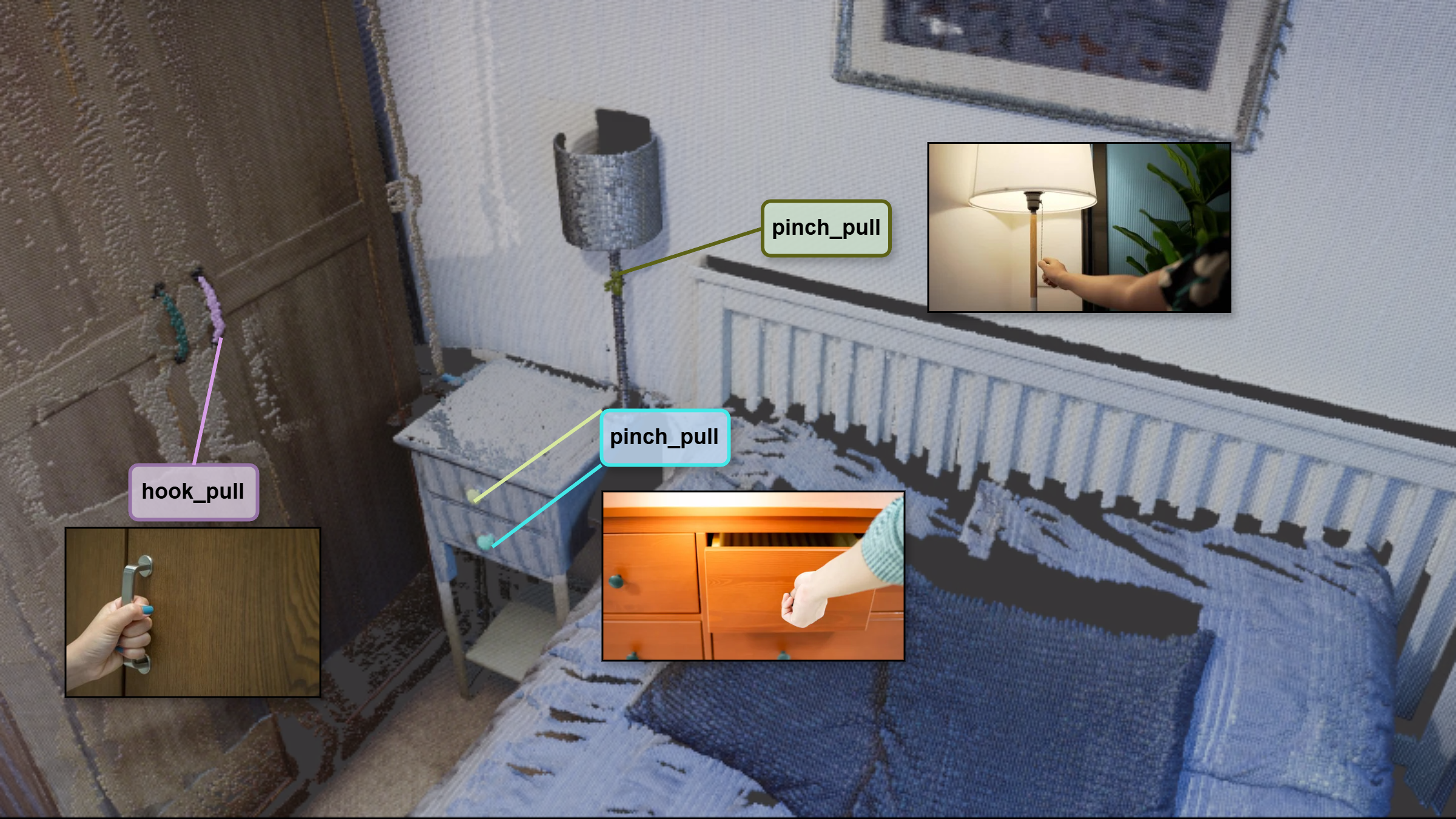}}&
\shortstack{\includegraphics[width=0.32\linewidth, height=0.19\linewidth]{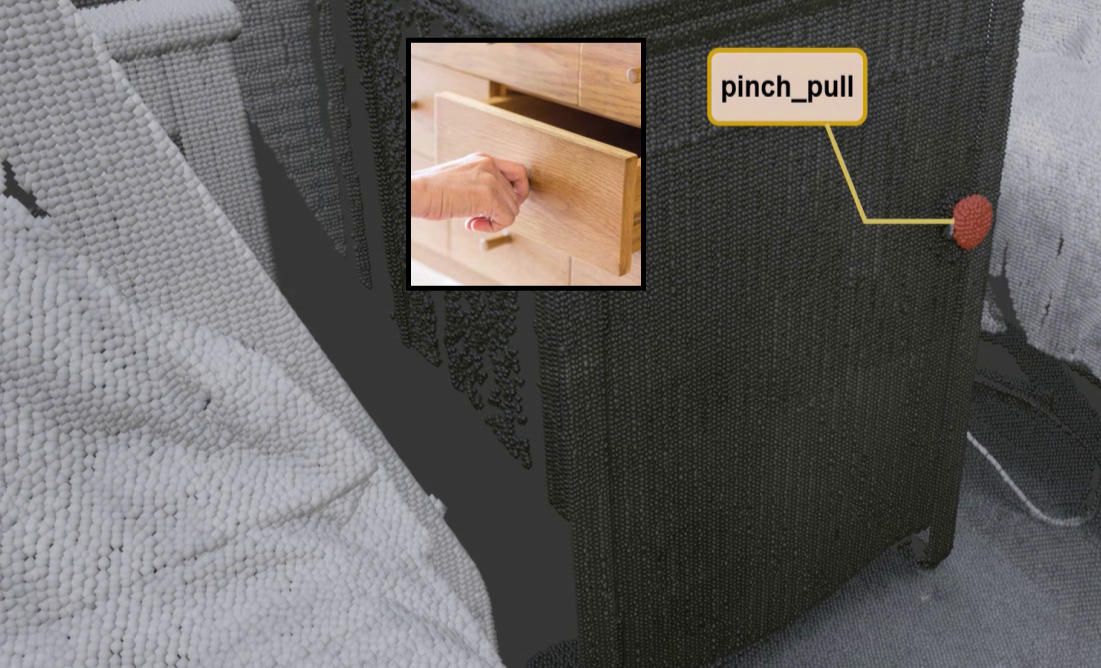}}&
\shortstack{\includegraphics[width=0.32\linewidth, height=0.19\linewidth]{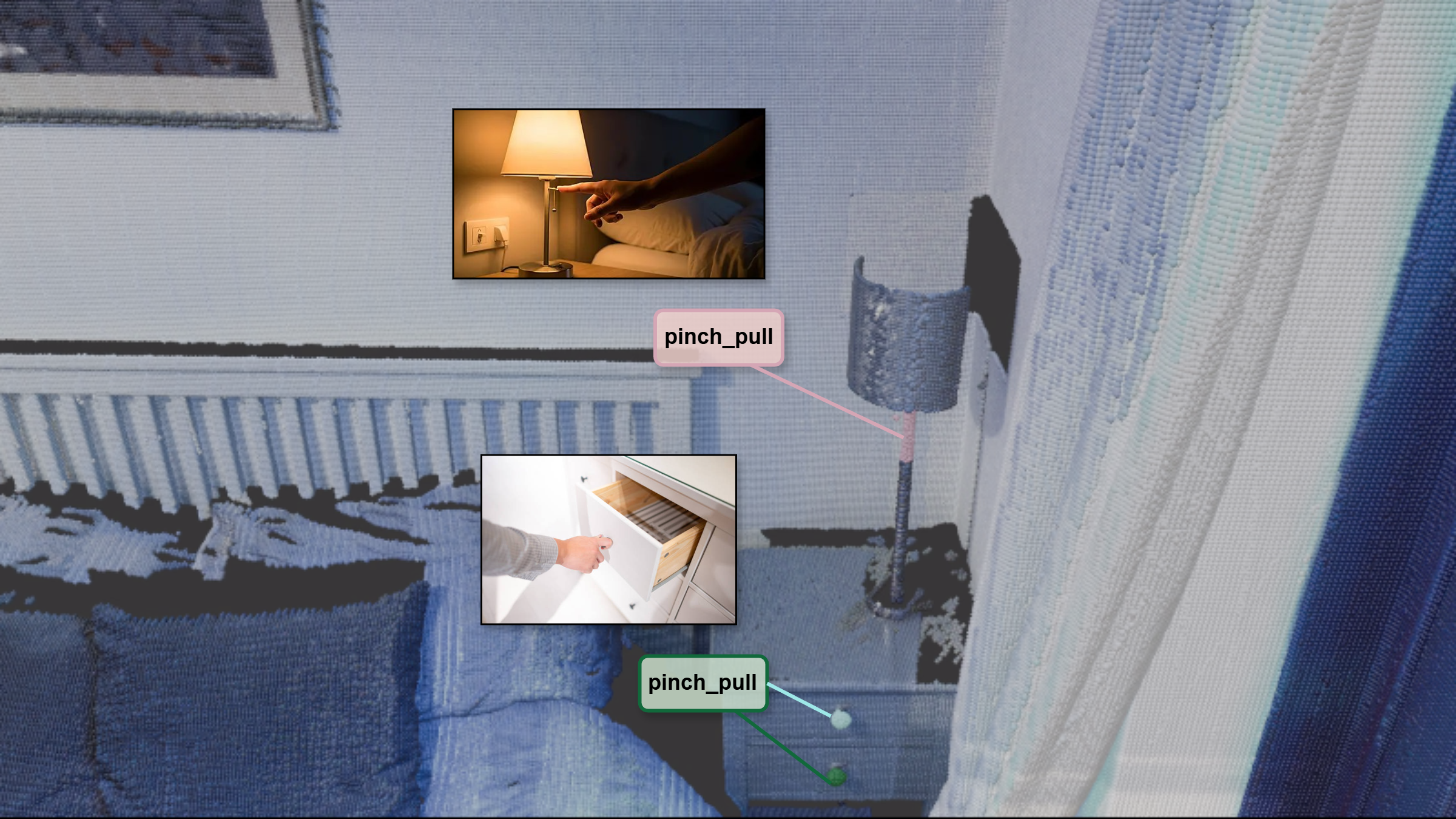}}
\end{tabular}
    \vspace{-1ex}
\captionof{figure}{Sample visualization of \textit{AffordBridge} dataset. 
} 
    \label{fig:result_Vis}
\end{center}
}]


\begin{abstract}
This supplementary material provides additional dataset analysis, empirical insights, and qualitative evaluations to complement our paper. We begin by providing more details about our \textit{AffordBridge} dataset, which features visualizations that illustrate affordance areas matched across multimodal inputs, including visual signifiers, textual descriptions, and action labels. We then discuss potential \textit{dataset usages}, highlighting research directions in 3D scene understanding and human-scene interaction enabled by our annotations. Next, we present results from our \textit{user study} evaluating the perceptual quality of our proposed \textit{AffordMatcher} against baselines, demonstrating superior correctness and interpretability. Finally, we visualize representative \textit{failure cases} of our method to underscore known limitations, such as handling sparse point clouds, complex actions, and ambiguous visuals, thereby reinforcing key observations and identifying opportunities for future work. Please see our video demonstration for a more interactive experience.

\end{abstract}

\section{Dataset Visualization} 
Fig.~\ref{fig:result_Vis} provides a sample visualization of the \textit{AffordBridge} dataset, highlighting affordance areas and their corresponding action phrases across different modalities. This figure presents an input scene alongside affordance areas matched to a visual signifier, demonstrating how environmental cues are linked to potential interactions.  For more details, please visit our demonstration video. 


\section{Potential Usages of Our Dataset} Our introduced \textit{AffordBridge} dataset includes annotations for both 3D scenes and RGB images to support affordance reasoning. Below, we outline several exciting research directions that can benefit from leveraging our dataset:
\begin{itemize}
    \item \textbf{3D Scene Understanding.} Traditional approaches to 3D scene analysis often focus on the instance level~\cite{sun2023superpoint, kolodiazhnyi2024oneformer3d}. By providing annotations for interactive elements on objects, our dataset opens opportunities for addressing various tasks in 3D scene understanding, such as 3D object detection and segmentation.
    \item \textbf{Robotic Manipulation.} Robots with affordance-aware systems can perform tasks more naturally, such as grasping~\cite{vuong2024language}, opening~\cite{wu2023learning}, or assembling objects in complex, unstructured environments~\cite{yamazaki2024open}. The release of our dataset could help robotic systems to understand the purpose and function of objects in a 3D context.
    \item \textbf{Human-Scene Interaction.} With affordance masks for 3D indoor scenes and bounding boxes for RGB images, researchers can gain deeper insights into the functional regions of objects and their interactions with humans. This can contribute to the development of more robust human-object interaction models that integrate 2D and 3D data~\cite{li20232d3d, li2020detailed}, facilitating unified interaction reasoning~\cite{linghu2024multi}.
\end{itemize}
\section{\textit{AffordMatcher} Analysis}
\textbf{User Study.} We conducted a user study to evaluate the perceptual quality of semantic affordance masks produced by our proposed \textit{AffordMatcher} versus three baselines, including Mask3D-F~\cite{delitzas2024scenefun3d}, PIAD~\cite{yang2023grounding}, and Ego-SAG~\cite{liu2024grounding}. Twenty experts in 3D vision each reviewed $40$ scenes ($10$ per method), rating the correctness of each affordance mask on a $5$-point Likert scale ($1$ = completely incorrect, $5$ = perfect) and selecting the single best segmentation per scene.

\begin{figure}[h]
   \centering
\includegraphics[width=\linewidth]{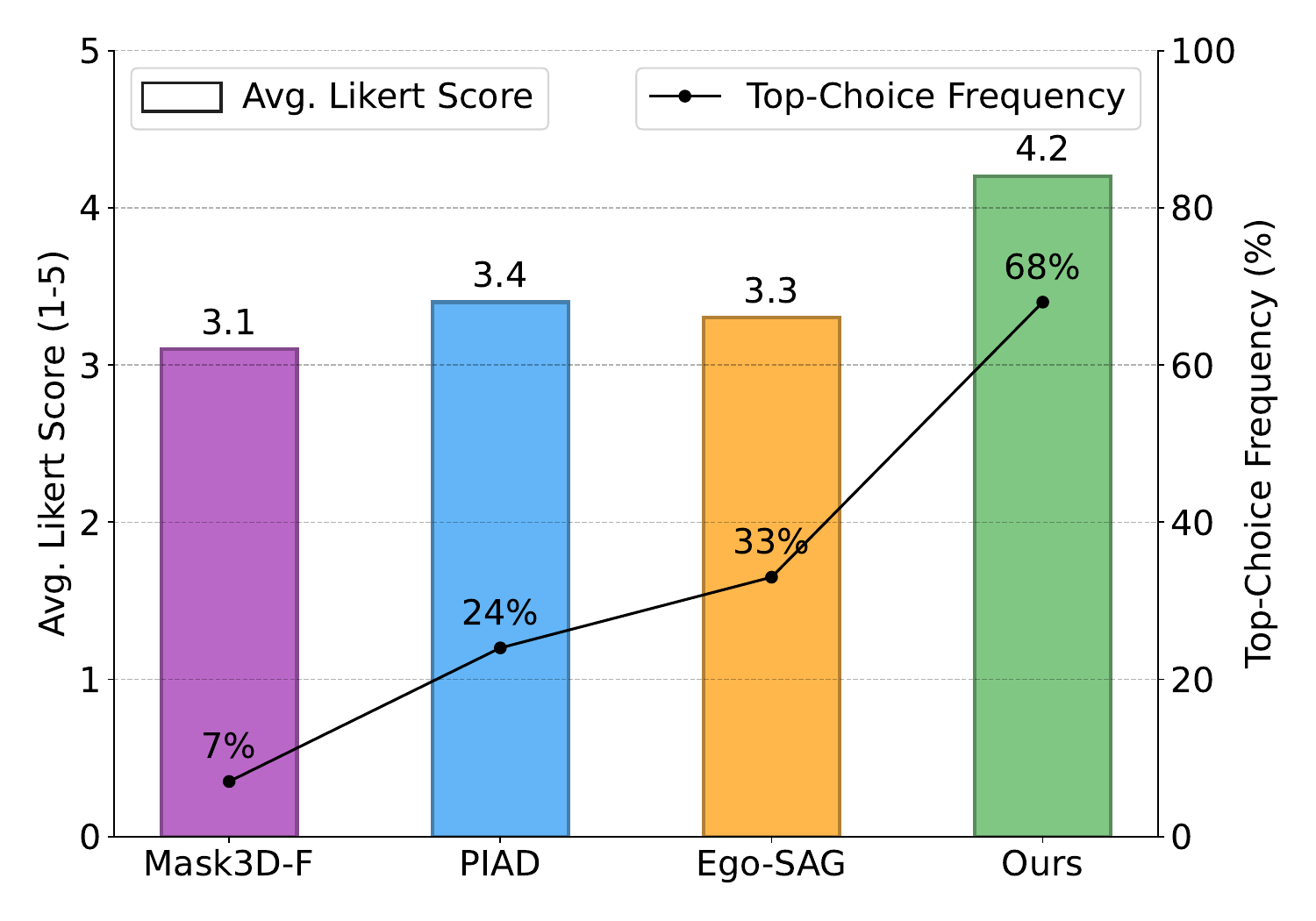}
  \vspace{-4ex}
\caption{User study results in interaction matching criteria.}
\vspace{-2ex}
\label{fig:userStudy}
\end{figure}
Fig.~\ref{fig:userStudy} presents the aggregated results: average Likert scores (bars) and the frequency each method was chosen as the top segmentation (line). Our approach achieved an average rating of $4.2$, substantially higher than Mask3D-F ($3.1$), Ego-SAG ($3.3$), and PIAD ($3.4$), and was selected as best in $68$\% of trials, significantly outperforming all baselines ($p < 0.01$, paired t-test).

Participants noted that our masks more accurately captured fine-grained affordance regions, such as chair seats or door handles, and avoided spurious activations common in baseline outputs. This confirms that our \textit{AffordMatcher} not only improves metric performance but also delivers actionable affordance in 3D point clouds.

\begin{figure}[h]
   \centering
\includegraphics[width=\linewidth]{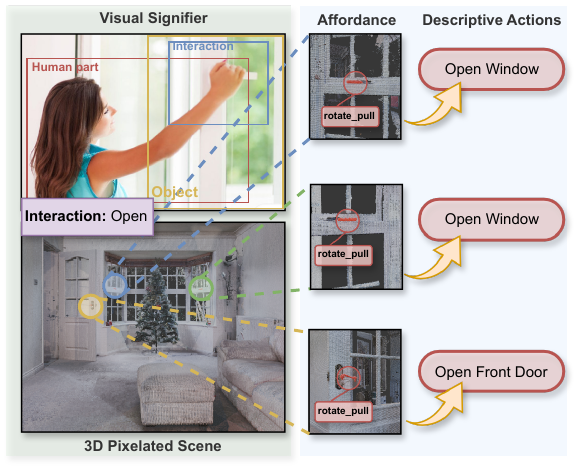}
\vspace{-4ex}
\captionof{figure}{Affordance prediction for a specific interaction, but results in different objects.
}
\label{fig:AffPredOverObj}
\vspace{-2ex}
\end{figure}

\begin{figure}[h]
  \centering
  \setlength\tabcolsep{5pt} 
  \renewcommand\arraystretch{0.6} 
  \begin{tabular}{cc}
    \begin{tabular}{@{}c@{}}
      \includegraphics[width=0.46\linewidth, height=0.32\linewidth]{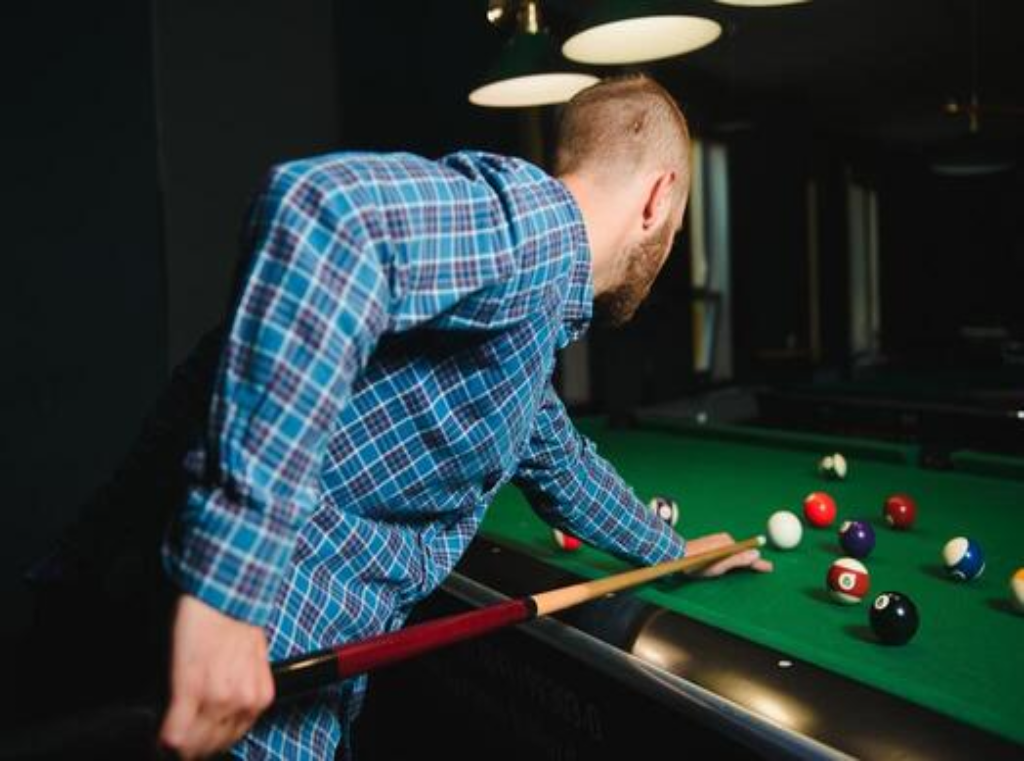} \\ \begin{tabular}{@{}c@{}}{Visual Signifier} \end{tabular}
      
    \end{tabular} 
    &
    \begin{tabular}{@{}c@{}}
      \includegraphics[width=0.46\linewidth, height=0.32\linewidth]{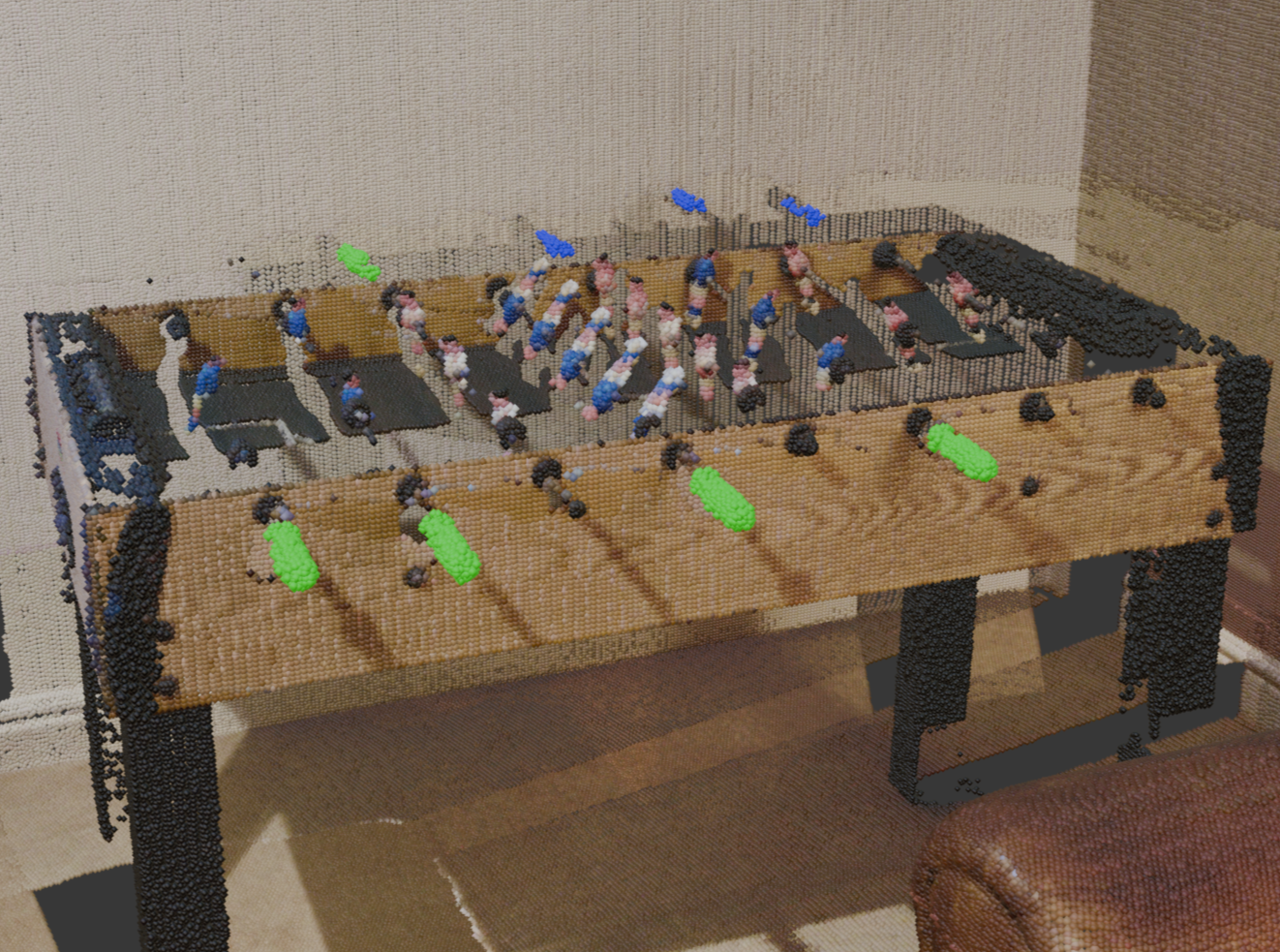} \\
      \begin{tabular}{@{}c@{}}{Output Affordances} \end{tabular}
    \end{tabular}\\
    \multicolumn{2}{c}{(a)}\\\\
    \begin{tabular}{@{}c@{}}
      \includegraphics[width=0.46\linewidth, height=0.32\linewidth]{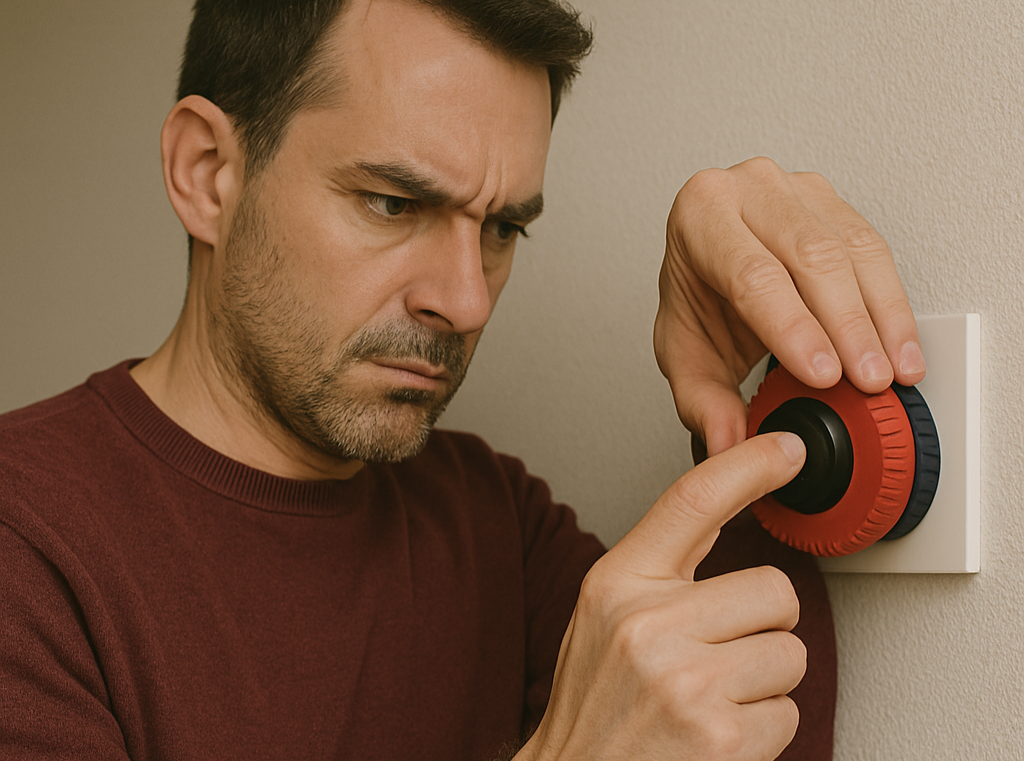} \\
      \begin{tabular}{@{}c@{}}{Visual Signifier} \end{tabular}
    \end{tabular}
    &
    \begin{tabular}{@{}c@{}}
      \includegraphics[width=0.46\linewidth, height=0.32\linewidth]{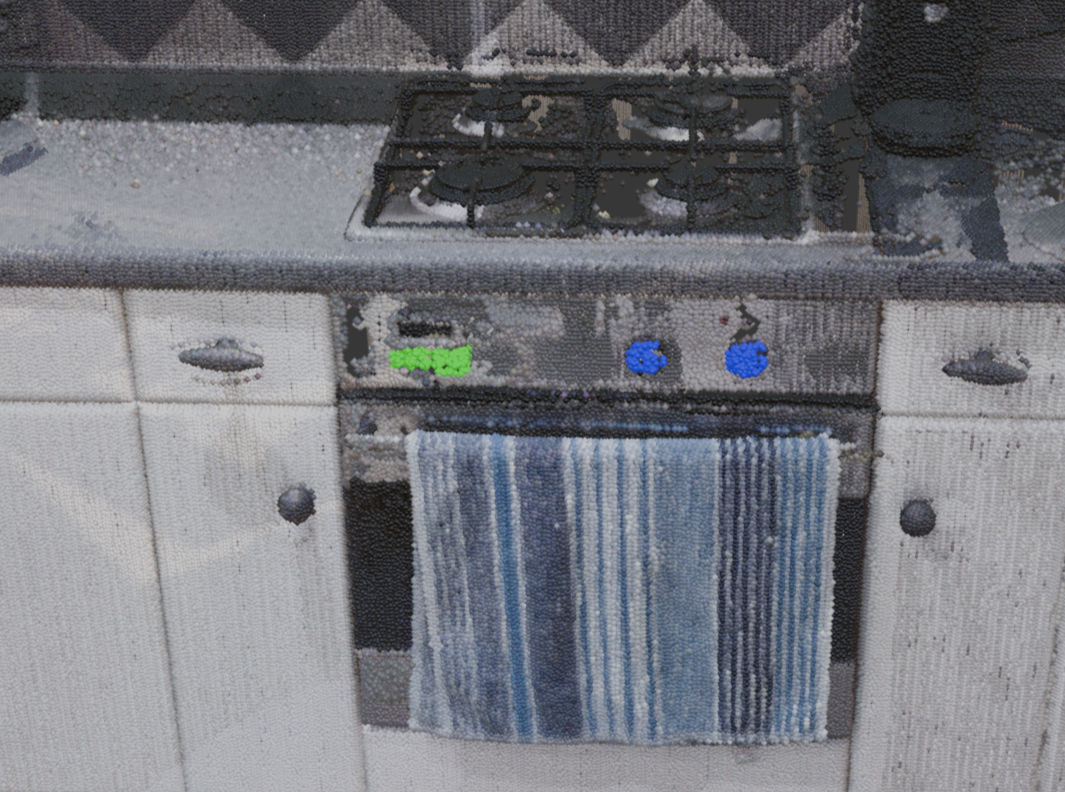}  \\
      \begin{tabular}{@{}c@{}}{Output Affordances} \end{tabular}
    \end{tabular}
    \\ \multicolumn{2}{c}{(b)}\\
  \end{tabular}
  \vspace{-2ex}
  \caption{Fail cases of our method: the visual signifier provides a complex action, which causes failure in reasoning. \textcolor{green}{\textbf{Green}} areas denote correct predictions; \textcolor{red}{\textbf{red}} and \textcolor{blue}{\textbf{blue}} areas are false positives and false negatives, respectively. } 
  \label{fig:failCase}
  \vspace{-4ex}
\end{figure}





\noindent\textbf{One-to-many Analysis.}
In Fig.~\ref{fig:AffPredOverObj}, we show the model's ability to localize the support regions required for a single interaction (“Open”) across multiple object instances within the same scene. From the top to bottom rows, we feed the network the same 2D visual cue-an outstretched hand poised to open-with the corresponding 3D voxelized indoor environment. The resulting affordance predictions correctly highlight the window latch, the second window’s handle, and finally the front door’s knob, each delineated by high-response voxels in the 3D scene. These results show that our \textit{AffordMatcher} flexibly generalizes the ``pitch\_pull" action to match with the interaction, successfully in semantically analogous parts on different objects, even when their appearance, scale, and orientation vary significantly.



\noindent\textbf{Fail Cases.} As outlined in the main paper, our method exhibits known limitations related to the challenge of semantic grounding, which are further exemplified through representative failure cases in Fig.~\ref{fig:failCase}. 
Specifically, such failure cases involve nuanced contextual cues or ambiguous object interactions that current models struggle to resolve without task-specific guidance. For example, in the Fig.~\ref{fig:failCase}\textcolor{red}{a}, the model failed to analyze the ``push" action when the man is playing billiards is from which side, or in the Fig.~\ref{fig:failCase}\textcolor{red}{b}, the model failed to distinguish whether the action in the visual signifier is the ``rotate" or the ``push".
Nonetheless, these examples serve to reinforce the overall generality of our approach under standard conditions, while motivating future work focused on enhancing semantic grounding and model adaptability in more challenging scenarios.

\end{document}